\documentclass[bookmarks=true]{article}

\usepackage[preprint]{corl_2026}
\usepackage{times}

\usepackage[numbers]{natbib}
\usepackage{multicol}
\usepackage{booktabs}
\usepackage{amsmath}
\usepackage{multirow}
\usepackage{graphicx}
\usepackage{caption}
\usepackage{placeins}
\usepackage{wrapfig}
\usepackage{subcaption}

\usepackage[activate={true,nocompatibility},final,tracking=true,kerning=true,spacing=true,factor=1100,stretch=10,shrink=10]{microtype}

\usepackage{amssymb}
\usepackage{tabularx} 

\usepackage{enumitem}
\usepackage{float}

\newcommand{\blue}[1]{\textcolor{blue}{{#1}}}
\newcommand{\algcmt}[1]{\hfill{\footnotesize\textcolor{blue}{// #1}}}
\usepackage{algorithm}
\usepackage{algpseudocode}


\begin{document}

% paper title
\title{PLanAR: Planning-Language-Grounded Agentic Reasoning for Robot Manipulation}

\author{
\small
\begin{tabular}{c}
\normalfont
Pengyuan Guo\textsuperscript{1,*},
Zhonghao Mai\textsuperscript{1,*},
Zhengtong Xu\textsuperscript{1,*},
Kaidi Zhang\textsuperscript{1},
Quan Khanh Luu\textsuperscript{1}
\\
\normalfont
Heng Zhang\textsuperscript{2},
Zichen Miao\textsuperscript{1},
Arash Ajoudani\textsuperscript{2},
Zachary Kingston\textsuperscript{1},
Qiang Qiu\textsuperscript{1},
Yu She\textsuperscript{1}
\\[2mm]
{\bfseries \textsuperscript{1}Purdue University
\qquad
\textsuperscript{2}Istituto Italiano di Tecnologia}
\\
{\bfseries \textsuperscript{*}Equal contribution}
\end{tabular}\\
{\small \blue{\url{https://planar-robot.github.io/}}}
}

\maketitle

\begin{center}
    \includegraphics[width=1\linewidth, trim=1.8cm 0.4cm 1.8cm 0.4cm, clip]{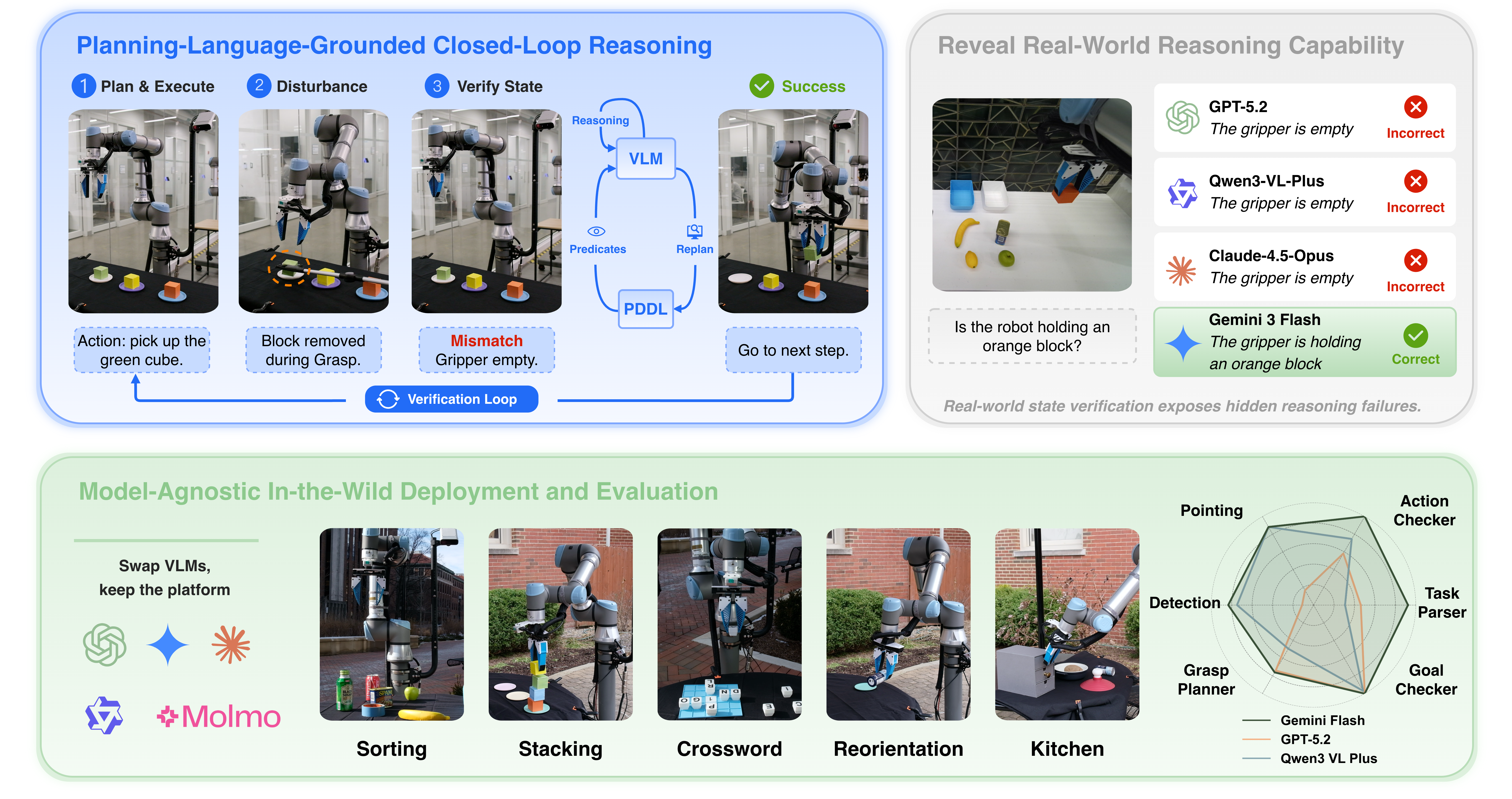}
    \captionof{figure}{PLanAR is a planning-language-grounded robot agent framework that enables vision-language models to perform structured agentic reasoning for long-horizon manipulation in unstructured real-world environments. With onboard perception and open-vocabulary prompts, PLanAR iteratively performs task planning, execution, online verification, and replanning.}
    \label{fig:teaser}
\end{center}
\vspace{+2.5em}

\begin{abstract}
Recent advances in vision-language models (VLMs) have enabled increasing progress in real-world robot manipulation. However, long-horizon manipulation in unstructured environments requires VLMs to reason about changing scene states, action constraints, and execution outcomes, which remains difficult with natural language reasoning alone. We present PLanAR, a planning-language-grounded robot agent framework for open-vocabulary, long-horizon manipulation. PLanAR uses a planning-language interface to define the VLM reasoning space: object predicates represent scene states, action schemas specify robot skills with preconditions and effects, and symbolic plans provide executable intermediate representations. This interface enables stepwise verification: after each action, PLanAR uses onboard observations to check whether the expected symbolic effects have been achieved, allowing the VLM-based agent to update task states, detect failures, and replan when execution deviates from expectation. Across robot embodiments, VLM backends, and tasks including stacking, crossword solving, and long-horizon kitchen workflows, PLanAR demonstrates strong real-world capability while revealing key limitations of current VLMs in embodied reasoning.
\end{abstract}


\section{Introduction}
Recent progress in large vision-language models (VLMs) has enabled strong open-vocabulary visual understanding \cite{achiam2023gpt,comanici2025gemini,yang2025qwen3}. These capabilities have motivated growing interest in using VLMs for real-robot manipulation \cite{intelligence2025pi_,team2025gemini,comanici2025gemini}. However, long-horizon manipulation in unstructured environments requires more than perception and one-step decision making. A robot must reason about changing scene states, action constraints, execution outcomes, and recovery as the world evolves \cite{team2025gemini,team2025gemini15}.

Existing methods are often limited by scene diversity, 
task complexity, or execution robustness: some operate 
in constrained settings with simplified object states \cite{zhang2024dkprompt,zhang2025llm}, others ground instructions in geometric 
representations without long-horizon symbolic reasoning \cite{huang2023voxposer,huang2024rekep,wu2023tidybot,liu2024ok}, and many rely on one-shot code generation or open-loop planning and execution without explicit verification and replanning \cite{ye2025pretraining,shen2026tiptop}. In parallel, vision-language-action models attempt to unify perception, reasoning, and control through end-to-end action prediction \cite{kim2024openvla,intelligence2025pi_,fang2026molmoact2,zha2026lap,intelligence2026pi07}, but strong performance often requires task-specific fine-tuning, which can reduce the open-world generalization inherited from pretrained models \cite{fei2025libero,zhou2025libero,kim2026molmospaces,yang2026robolab,shi2025hi}. 

A key challenge in using VLMs for manipulation is that natural language reasoning alone does not provide a reliable interface for long-horizon robot execution. Without an explicit representation of what 
states are relevant, which skills are admissible, and 
what each skill is expected to change, the agent can lose 
consistency across steps, hallucinate object states, or 
fail to recognize when replanning is needed. These issues compound under real-world occlusion, disturbances, and 
partial observability.

Planning languages provide a natural way to make long-horizon robot reasoning explicit and verifiable \cite{liu2023llmp, garrett2020pddlstream,kwon2025fast}. In classical planning, the Planning Domain Definition Language (PDDL) defines a domain through object predicates, action schemas, preconditions, and effects \cite{garrett2021integrated,helmert2006fast}. For VLM-based robot agents, this representation can serve as an explicit reasoning space: predicates make scene states explicit, action schemas constrain the agent to admissible robot skills, and preconditions and effects expose the causal structure of each action. Moreover, expected symbolic effects provide a direct mechanism for closed-loop verification. After executing an action, the agent can verify from onboard observations whether the intended state transition has been achieved and replan when needed. In this way, planning-language grounding supports not only task decomposition, but also online verification, failure recovery, and disturbance handling.

In this paper, we present \textbf{PLanAR}, planning-language-grounded agentic reasoning for robot manipulation. Our contributions are threefold.

1. \textbf{Planning-Language-Grounded Closed-Loop Reasoning.} PLanAR uses explicit object 
predicates, action schemas, preconditions, and effects. 
Since each action has expected symbolic effects, PLanAR 
can verify outcomes, update state, detect 
failures, and replan when necessary.

2. \textbf{Compositional Agentic Pipeline.} We instantiate this formulation as a compositional robot agent pipeline that decomposes long-horizon manipulation into perception, symbolic state estimation, task planning, skill execution, outcome verification, and replanning. This modular design connects VLM reasoning with executable robot skills while keeping each stage inspectable and replaceable, enabling systematic analysis of where embodied reasoning succeeds or fails.

3. \textbf{Generalist Real-World Deployment.} We build PLanAR as a real-world robot agent platform for in-the-wild manipulation. PLanAR supports different VLM backends and robot embodiments, enabling model-agnostic, cross-embodiment deployment on long-horizon tasks such as spatially constrained stacking, crossword solving, and kitchen workflows.

\section{Related Work}

\subsection{VLM-based Robot Manipulation}

Recent progress in VLMs has motivated their use as planning and reasoning modules for robot manipulation. A common paradigm is to condition a VLM on visual observations and language instructions to generate plans \cite{liu2024moka,ye2025pretraining,chiang2024mobility,shen2026tiptop,duan2024manipulate} or programs \cite{huang2023voxposer,huang2024rekep,fu2026capx,jin2024robotgpt}, which are then translated into executable skill calls handled by separate perception and control modules. While promising, these systems are often constrained by curated scenes \cite{zhang2024dkprompt,huang2023voxposer}, simplified object states \cite{ye2025pretraining,jin2024robotgpt,zhang2025llm,gong2025anytask,athalye2026pixelspredicateslearningsymbolic}, and short-horizon tasks \cite{shen2026tiptop,liu2024moka,huang2024rekep}.

Some work grounds language instructions in spatial or geometric representations, such as object-centric constraints, affordance maps, keypoints, or spatial value functions \cite{huang2023voxposer,huang2024rekep,wu2023tidybot,liu2024ok,fang2025saga,patel2025real,kumar2026openworldtaskmotionplanning}. These methods are effective for geometry-dominated manipulation, but place less emphasis on semantic long-horizon reasoning over evolving symbolic states and execution outcomes.

Closed-loop execution remains challenging. Many VLM-based systems rely on one-shot code generation, open-loop planning, or limited feedback after execution \cite{ye2025pretraining,shen2026tiptop}. Without explicit expected effects for each action, it is difficult to verify success, update task states, or decide when to replan. Existing task-specific checks or corrective behaviors \cite{fu2026capx,duan2024manipulate} are often heuristic rather than grounded in a unified reasoning representation, making long-horizon execution brittle under object movement, occlusion, grasp failures, disturbances, and background changes.

PLanAR addresses these limitations by grounding VLM reasoning in a planning-language interface for manipulation. Object predicates, action schemas, preconditions, and effects define an explicit reasoning space, while expected symbolic effects provide natural verification targets. This enables closed-loop agentic reasoning: the agent checks execution outcomes from onboard observations, and replans when the scene deviates from expectation.

\subsection{Embodied Benchmarks for VLMs}

Many embodied benchmarks for VLMs use question-answering or offline interfaces to evaluate visual and spatial language reasoning without executing actions in the world \cite{team2025gemini,song2024robospatial,majumdar2024openeqa,cheng2024spatialrgpt,team2025gemini15,hong2026esibench}. While useful for controlled diagnosis, these settings cannot test whether a model can maintain grounding under action-induced scene changes, verify execution outcomes, recover from failures, or handle error accumulation over long horizons.

Simulation-based benchmarks provide scale through standardized task suites \cite{yang2025embodiedbench,li2024embodied,khanna2024goat,liu2023agentbench,choi2024lota,liu2024visualagentbench,kim2026molmospaces,yang2026robolab,huang2026kinderphysicalreasoningbenchmark}. However, simulated environments often simplify physical dynamics, object states, perception noise, and action execution through cleaner scenes or abstracted primitives. As a result, simulation success may overestimate robustness and does not always translate to reliable real-world manipulation.

PLanAR provides real-world, closed-loop evaluation of VLM agents on physical robot tasks, revealing embodied reasoning limitations under clutter, occlusion, lighting variation, object movement, and scene changes that offline and simulation benchmarks often miss.

\section{PLanAR Framework}
\begin{figure*}[t]
  \centering
  \includegraphics[width=1\textwidth,  trim=0 0 0 0, clip]{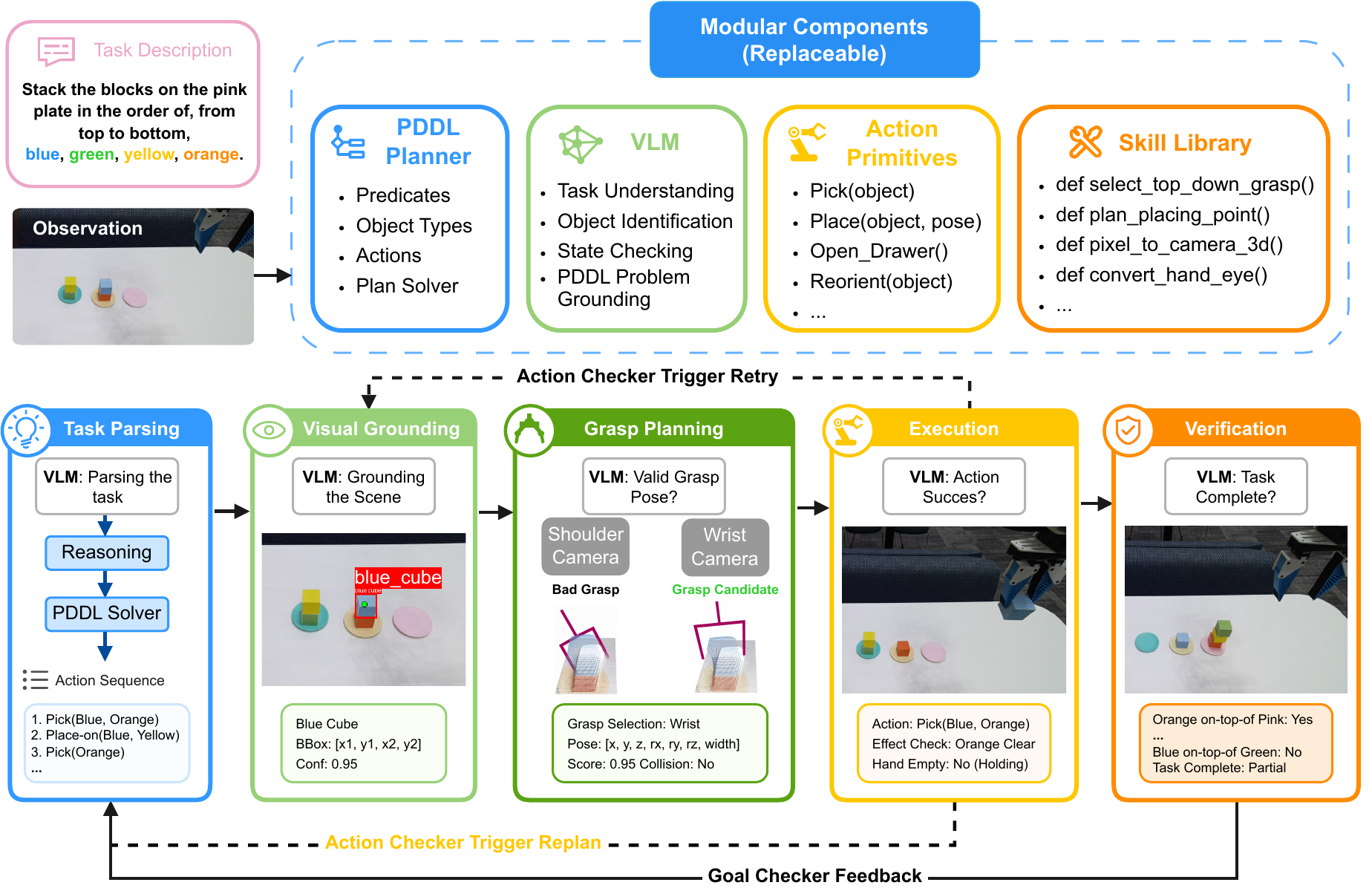}
  
  \caption{Pipeline of PLanAR. Given a natural-language instruction and onboard observations, the task parser grounds the instruction into symbolic goals and a PDDL action sequence. Visual grounding and grasp planning then instantiate objects, states, and feasible grasps in the real scene. Action and goal checkers verify states after each step; failed checks trigger per-step retry or replanning with updated observations. All modules are replaceable under the same planning-language interface.}
  \label{fig:pipeline_overview}
\end{figure*}

As shown in Fig.~\ref{fig:pipeline_overview}, PLanAR connects natural-language task instructions, onboard observations, symbolic planning, robot skill execution, and closed-loop verification within a unified pipeline. To bridge unstructured language instructions and executable robot logic, PLanAR grounds VLM reasoning in a PDDL interface, building on LLM-guided symbolic planning formulations~\cite{liu2023llmp, zhang2024dkprompt} and extending them to closed-loop verification and replanning. The PDDL domain defines the agent's symbolic reasoning space through object types, predicates, action schemas, preconditions, and effects, such as whether an object is on a support surface or held by the gripper. Given a generated PDDL problem and the predefined domain, PLanAR invokes Fast Downward~\cite{helmert2006fast} to compute a sequence of high-level symbolic actions that satisfy the task goal. The resulting plan provides an executable symbolic interface between VLM reasoning and robot control. Implementation details are provided in Appendix~\ref{app:Implementation_details}.

\subsection{Symbolic Verification and Replanning}
Before each action, the action checker evaluates whether the required preconditions hold and whether the action is feasible in the current scene. After execution, the system re-observes the environment and verifies that the expected effects have taken place, using updated visual input to detect failures such as missed grasps, incorrect object interactions, or unexpected state changes. Upon completion of the full action sequence, a final goal-condition check confirms whether the scene satisfies all specified goal predicates. When a verification step fails, PLanAR triggers replanning: the task parser re-instantiates an updated PDDL problem reflecting the current world state, and the symbolic planner generates a revised action sequence from that state. This closed loop between visual observation, symbolic verification, and plan repair enables robust long-horizon manipulation without assuming perfect execution at each step.

\subsection{Converting Symbolic Actions into Robot Skills}
To bridge the gap between discrete symbolic plans and continuous physical execution, PLanAR abstracts robot behavior into a library of high-level parameterized action primitives, including pick, place, and open/close drawer. To execute these primitives, PLanAR leverages specialized skill library functions, such as coordinate transformations and placement pose generation, extending the system's capabilities beyond basic pick-and-place to include complex interactions like bottle reorientation, drawer manipulation, and constraint-driven stacking. Visual grounding uses a swappable interface supporting direct VLM prediction or LangSAM~\cite{medeiros2023langsam} segmentation, enabling open-vocabulary detection across models with or without native grounding capability. For grasping actions, PLanAR first grounds the target object from RGB-D observations and generates candidate 6-DoF grasp poses using AnyGrasp~\cite{fang2023robust}. A VLM-based grasp evaluator then checks whether the selected grasp is semantically correct and physically feasible, including whether it targets the intended object and whether it may cause collision or instability. As shown in Fig.~\ref{fig:pipeline_overview}, PLanAR first plans grasps from the shoulder-view camera. If the candidate grasp is rejected, the robot switches to the wrist-mounted camera for close-range observation and replans the grasp with more localized visual evidence.

\begin{wrapfigure}{r}{0.5\linewidth}
  \centering
  \vspace{-2em}
  \includegraphics[width=\linewidth, trim=20 40 17 10, clip]{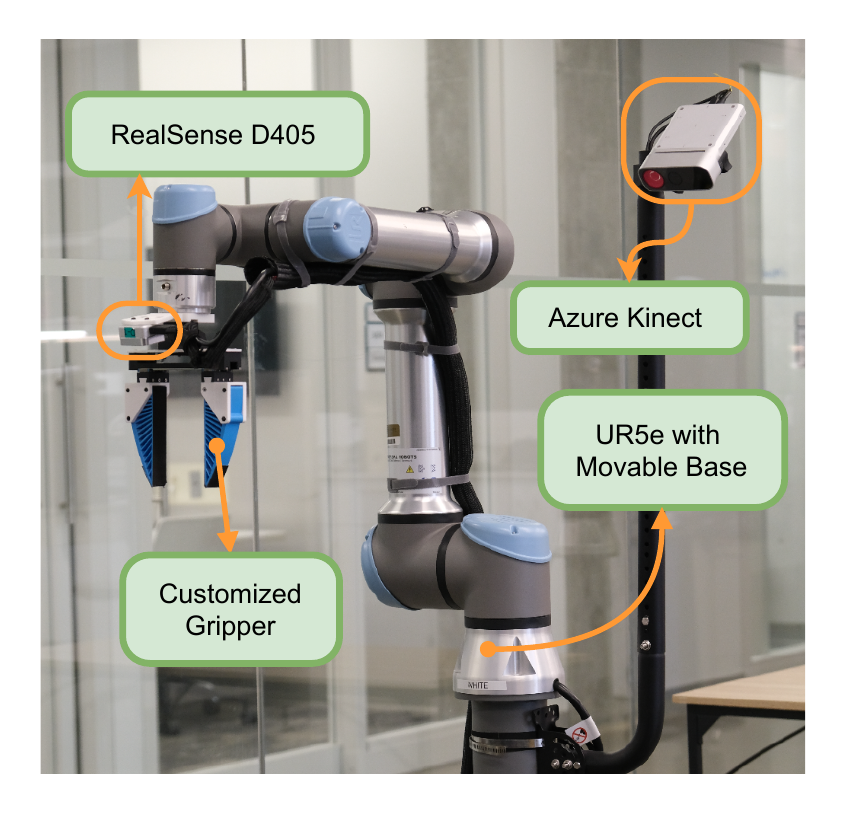}
  \caption{The PLanAR Manipulation Platform: a reproducible, easy-to-deploy platform for real-world in-the-wild manipulation.} 
  \label{fig:hardware_setup}
  \vspace{-3em}
\end{wrapfigure}

\subsection{Implementation Across Robot Embodiments}
We instantiate PLanAR on a UR5e platform with two RGB-D views: a fixed shoulder-view camera for global scene grounding and a wrist-mounted camera for close-range grasp verification and replanning. The UR5e is mounted on a movable base for deployment across laboratory settings, general indoor environments, and in-the-wild scenes. It uses a low-cost parallel gripper with fin-ray fingers for repeatable manipulation of diverse everyday objects. Fig.~\ref{fig:hardware_setup} provides an overview of the platform. Because PLanAR exposes a structured interface, models and robot embodiments can be swapped under the same action schema and verification protocol. We also transfer PLanAR to a Franka Panda arm using the same symbolic domain, action schemas, and verification prompts, changing only the low-level execution wrapper. 

\section{Experiment}

We organize the experiments around four questions. First, how reliably do VLM-based robot agents perform closed-loop manipulation? Second, can PLanAR's modular interface mitigate specific bottlenecks by assigning different VLMs to different modules? Third, how much do planning-language grounding and execution-time verification contribute to robustness? Finally, can PLanAR transfer across robot embodiments and compare favorably against VLA-based baselines?

We validate PLanAR on two real robot platforms. On the UR5e, we evaluate five tasks, {Sorting}, {Stacking}, {Crossword}, {Reorientation}, and {Kitchen}, across laboratory, indoor, and outdoor settings, as shown in Fig.~\ref{fig:task}. On Franka Panda, we evaluate Sorting and Stacking for cross-embodiment comparison, including direct VLM planning and fine-tuned VLA baselines. These tasks test open-vocabulary grounding, spatial reasoning, symbolic state tracking, and recovery under real-world uncertainty.

\subsection{Diagnosing VLM Bottlenecks in Closed-Loop Manipulation}
\label{sec:e2e_failure_analysis}
We first evaluate eight single-VLM pipelines on the sorting task, using 20 real-robot trials across six object setups spanning fruits, blocks, toys, and tools. Fig.~\ref{fig:failure_modes} summarizes dominant failure modes; full rollout statistics are reported in Appendix Table~\ref{tab:single_vlm_end2end}.

Gemini Flash is the most reliable single-backend pipeline, combining low latency with stable visual grounding. Gemini Pro is strong in abstract reasoning but less reliable in visual grounding; replacing its detector with LangSAM improves performance.

\begin{figure*}[t]
  \centering
  % Replace with your actual filename
  \includegraphics[width=\linewidth]{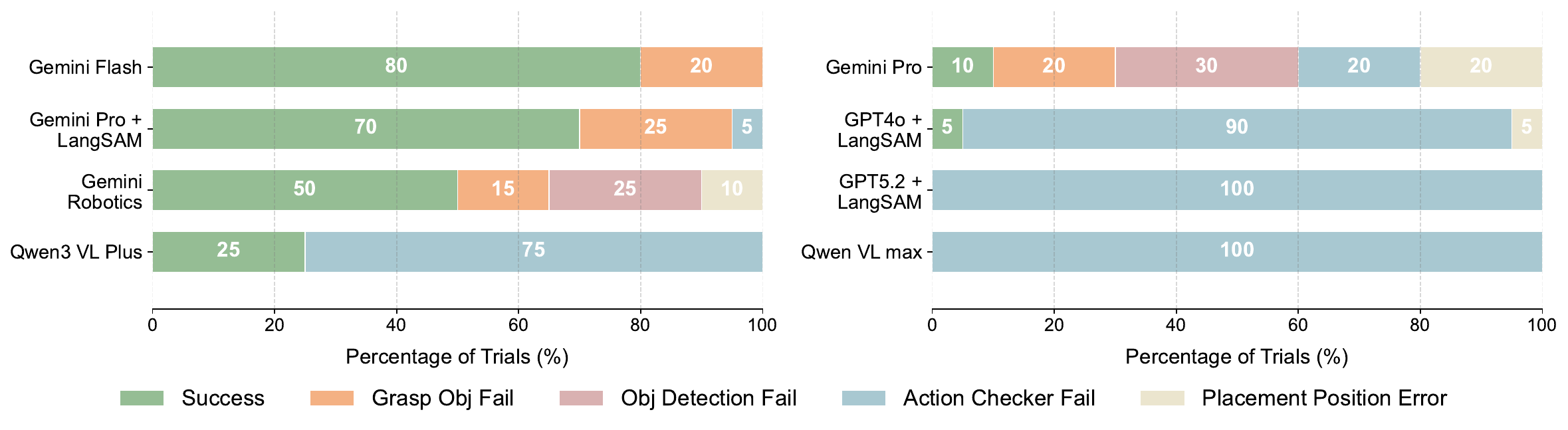}
  
  \caption{Failure mode breakdown for single-VLM pipelines on the sorting task. Results show which pipeline components dominate failures across different VLMs.}
  \label{fig:failure_modes}
\end{figure*}

Models such as Qwen-VL-Max and GPT-5.2 achieve near-zero success primarily due to unreliable execution-state verification. A common failure mode is hallucination during action checking. For example, the model may claim the object is not in the gripper after a successful pick.

To attribute these rollout failures to concrete sources, we further evaluate five modules in the PLanAR pipeline: task parser, object detector, action checker, grasp planner, and goal checker, with full results reported in Table~\ref{tab:module_breakdown}. Task parsing is generally not the main bottleneck for cloud-hosted VLMs. Instead, failures are dominated by grounding and verification, whose errors compound because they are repeatedly invoked during execution. For example, in our sorting task, placing three objects triggers six calls to the action checker. As a result, even with 90\% accuracy on static VQA, the probability of completing a full trial is only $0.9^6 \approx 0.53$. More critically, at 50\% accuracy, success collapses to $0.5^6 \approx 1.5\%$. This compounding effect explains why models with seemingly moderate verification errors in isolation can fail catastrophically under closed-loop execution.

\begin{table*}[t]
\centering
\caption{Module evaluation of VLMs for PLanAR. SR denotes success rate; ``--'' indicates the VLM cannot generate usable results for the module.}
\label{tab:module_breakdown}
\setlength{\tabcolsep}{4pt}
\resizebox{\linewidth}{!}{%
\begin{tabular}{ll | c c | c c | cc c | c c | c c}
\toprule
\multirow{2}{*}{\textbf{Model}} &
& \multicolumn{2}{c|}{\textbf{Task Parser}}
& \multicolumn{2}{c|}{\textbf{Action Checker}}
& \multicolumn{3}{c|}{\textbf{Object Detector}}
& \multicolumn{2}{c|}{\textbf{Grasp Planner}}
& \multicolumn{2}{c}{\textbf{Goal Checker}} \\
\cmidrule(lr){3-4} \cmidrule(lr){5-6} \cmidrule(lr){7-9} \cmidrule(lr){10-11} \cmidrule(lr){12-13}
& & \textbf{SR (\%)} & \textbf{Time}
  & \textbf{SR (\%)} & \textbf{Time}
  & \textbf{Pointing~SR (\%)} & \textbf{Bbox~IoU (\%)} & \textbf{Time}
  & \textbf{SR (\%)} & \textbf{Time}
  & \textbf{SR (\%)} & \textbf{Time} \\
\midrule

% Gemini
\multirow{3}{*}{\rotatebox{90}{Gemini}}
& Gemini 3 Pro    & \textbf{92.3} & 70.5 & 98.4 & 13.3 & 66.7  & 53.7 & 45.7 & \textbf{77.8} & 26.8 & \textbf{100.0} & 8.4 \\
& Gemini 3 Flash  & \textbf{92.3} & 45.2 & \textbf{100.0} & \textbf{5.3} & \textbf{88.1} & \textbf{83.4} & 9.5 & 74.1 & 9.1 & \textbf{100.0} & 15.6 \\
& Gemini Robotics & 75.0 & 29.6 & 95.1 & 10.1 & 57.1 & 40.4 & 5.8 & 55.0 & 10.3 & \textbf{100.0} & 13.5 \\
\midrule

% Qwen
\multirow{2}{*}{\rotatebox{90}{Qwen}}
& Qwen3-VL-Plus & 30.8 & 40.2 & 74.6 & 11.4 & \textbf{88.1} & 74.8 & 5.1 & 66.7 & 10.0 & 96.9 & 12.1 \\
& Qwen-VL-Max   & 17.3 & 36.3 & 64.8 & 24.2 & 81.0 & 60.3 & 5.0  & 59.2 & 9.7 & 84.4 & 11.2 \\
\midrule

% Others
\multirow{4}{*}{\rotatebox{90}{Others}}
& GPT-5.2           & 46.2 & 31.8 & 58.2 & 8.6 & 16.6 & 11.2 & \textbf{3.2} & 75.9 & 8.0 & \textbf{100.0} & 9.1 \\
& GPT-4o            & 48.1 & \textbf{16.0} & 68.9 & 7.0 & 14.3 & 8.0  & 3.9 & 44.4 & \textbf{4.4} & 93.8 & \textbf{5.1} \\
& Claude-4.5-Opus   & 78.9 & 17.0 & 64.8 & 7.1 & 9.6  & 6.1  & 4.4 & 70.4 & 8.2 & 93.8 & 6.6 \\
& Claude-4.5-Sonnet & 40.4 & 17.2 & 40.4 & 11.9 & 61.9 & 43.9 & 4.1 & 54.7 & 9.6 & 78.1 & 10.0 \\
\midrule

% Local
\multirow{2}{*}{\rotatebox{90}{Local}}
& Qwen2.5-VL-7B & -- & -- & 25.8 & 8.1 & 73.8 & 51.0 & 6.6 & -- & -- & 50.0 & 9.2 \\
& Molmo-7B   & -- & -- & 33.9 & 19.0 & 76.2 & 27.0 & 3.6 & -- & -- & 43.8 & 18.1 \\
\bottomrule
\end{tabular}}
\end{table*}

\subsection{Modular VLM Composition} 
\label{sec:composite}

\begin{wrapfigure}{r}{0.5\linewidth}
  \centering
  % Replace with your actual filename
  \includegraphics[width=\linewidth, trim=5 0 10 20, clip]{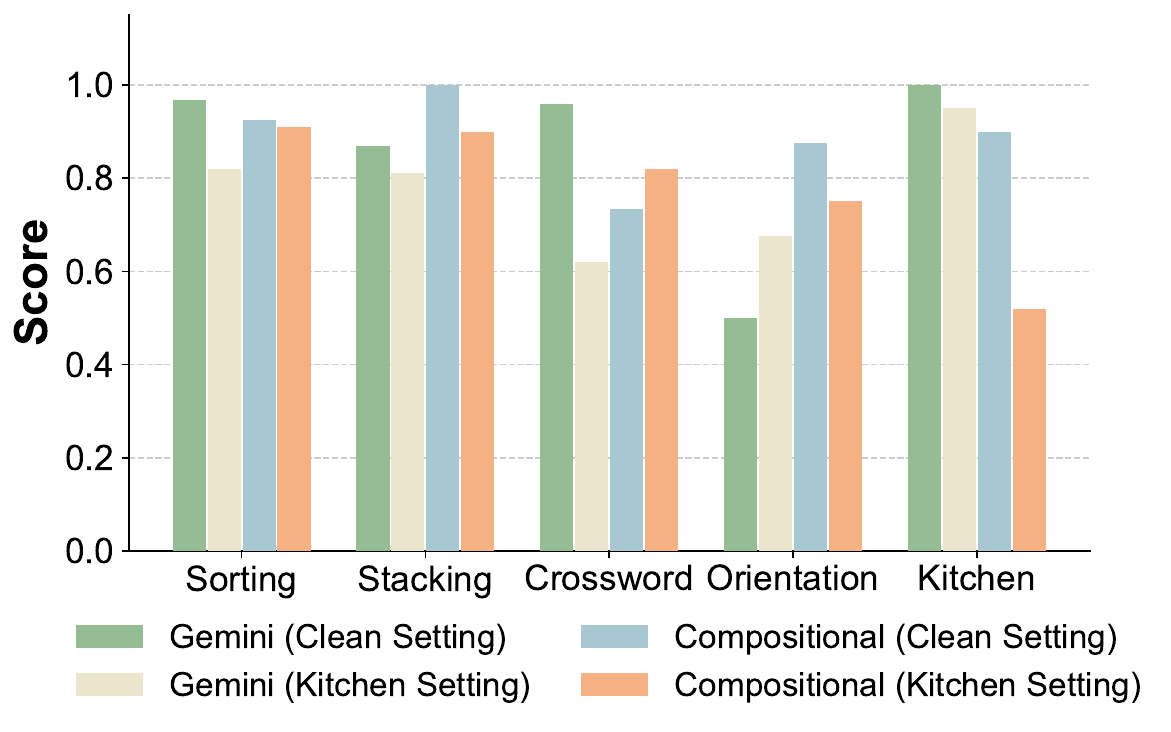}
  \vspace{-1em}
  \caption{Performance comparison across environments. We compare a Gemini Flash baseline with a compositional pipeline across five manipulation tasks. Performance is measured using a task progress score that quantifies partial completion.}
  \label{fig:task_performance_chart}
  \vspace{-1.5em}
\end{wrapfigure}

Motivated by the module-level evaluation, we next ask whether PLanAR's modular interface can mitigate specific bottlenecks through model assignment. We construct a compositional pipeline that assigns different VLM backends to different modules: Gemini Flash for the task parser and grasp planner, Qwen3-VL-Plus for the object detector, and Claude Opus for the goal checker. These models are selected based on their competitive module-level performance. We compare this compositional pipeline against a strong single-backend Gemini Flash baseline across laboratory and kitchen settings, with results reported in Fig.~\ref{fig:task_performance_chart}.

For this experiment, we report a task progress score computed from a structured rubric in Appendix~\ref{app:task_setup_metrics}. This metric captures partial completion and is more informative than binary success. To reduce rollout latency in longer tasks, we disable action-level verification and retain only the goal checker. We analyze the cost--benefit of dense action verification separately in Appendix~\ref{app:closed_loop_ablation}.

As shown in Fig.~\ref{fig:task_performance_chart}, the compositional pipeline does not consistently outperform the Gemini Flash baseline. Both pipelines show a robustness gap when moving from controlled laboratory trials to kitchen settings, with the largest drops in the Crossword and Kitchen tasks, where precise placement and sustained object-identity tracking are critical.

This suggests that model composition is not universally better than a strong single-VLM pipeline; its benefit depends on the task and scene. For example, replacing the grounding module with Qwen3-VL-Plus improves stacking, where localization and precise placement are important. In the kitchen setting, however, the same substitution exposes Qwen's spatial-grounding weakness, such as confusing the middle drawer with the top drawer, and performs worse than Gemini Flash. For the reorientation task, we use a task-specific Gemini 3 Pro variant for pose-conditioned reasoning, as most models struggle to jointly reason about the grasped bottle's orientation and its spatial relationship to the table. Overall, we use composition as a targeted strategy to exploit model complementarity when the bottleneck is identifiable, rather than as a universally superior design.

\subsection{Ablation: Planning-Language Grounding and Closed-Loop Verification}
\label{sec:pddl_verification_baseline}

We ablate the roles of planning-language grounding and execution-time verification on our Franka Panda adaptation of PLanAR. All configurations use the same VLM backend, perception stack, robot platform, and skill library: \textbf{VLM-Direct}, which directly predicts an action sequence; \textbf{VLM+PDDL (Open-Loop)}, which executes a symbolic plan without verification; and \textbf{PLanAR}, which adds step-wise precondition/effect verification and replanning.

We evaluate all configurations on fruit sorting, where the robot must place three fruits into a target bowl. To test robustness under execution uncertainty, a human removes the second grasped fruit and returns it to the table, creating a state mismatch similar to misgrasps, object slips, or failed placements.

\begin{figure}[t]
  \centering
  \includegraphics[width=\linewidth, trim=0 0 0 2, clip]{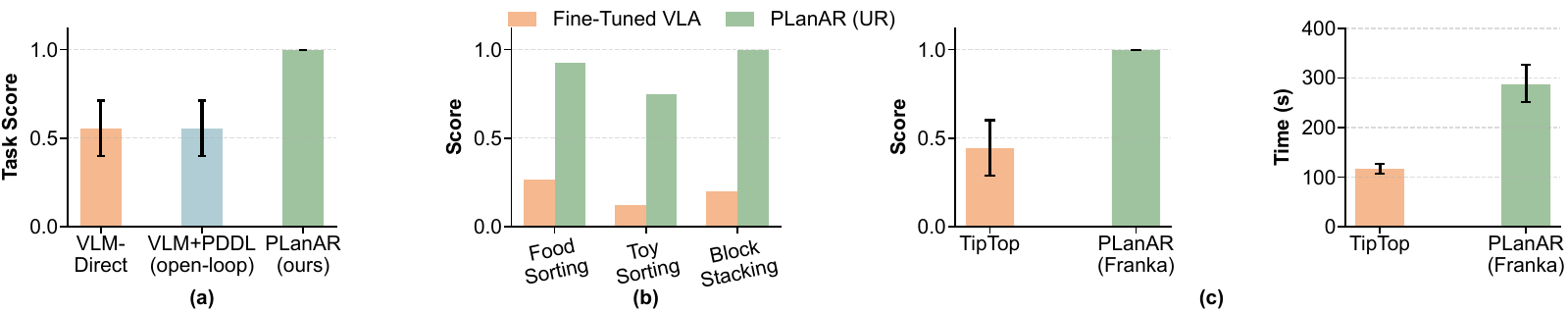}
  \caption{
  Ablations and baseline comparisons.
\textbf{(a)} Comparison of VLM-Direct, VLM+PDDL (Open-Loop), and PLanAR on fruit sorting with human disturbance.
\textbf{(b)} Comparison between PLanAR and a fine-tuned $\pi_{0.5}$ VLA baseline.
\textbf{(c)} Cross-embodiment evaluation on the Franka robot and comparison with TiPToP on fruit sorting with human disturbance.
  }
  \label{fig:pipeline_ablation}
\end{figure}

As shown in Fig.~\ref{fig:pipeline_ablation}(a), VLM-Direct and VLM+PDDL (Open-Loop) exhibit similar performance: both continue execution without detecting that the second grasp no longer holds. VLM-Direct lacks an explicit intermediate state representation for checking execution outcomes; VLM+PDDL provides a symbolic plan, but without execution-time verification it still proceeds under an invalid state assumption. In contrast, PLanAR detects that the expected grasp effect failed, updates the symbolic state, replans, and completes the task. This result shows that planning-language grounding structures action generation, but robustness comes from verifying symbolic effects during execution.

\subsection{Comparison with VLA Baselines and Cross-Embodiment Generalization}
\label{sec:vla}
Finally, we evaluate PLanAR against VLA-based baselines and examine whether its planning-language-grounded design transfers across robot embodiments. We first compare PLanAR with a fine-tuned $\pi_{0.5}$ baseline~\cite{intelligence2025pi_}. The $\pi_{0.5}$ policy is fine-tuned with 40 demonstrations for sorting and 30 demonstrations for stacking. We then compare PLanAR with TiPToP~\cite{shen2026tiptop} on a Franka Panda. Across these experiments, PLanAR uses the same high-level task interface, perception stack, planning-language grounding, and verification logic, while only adapting the low-level execution wrappers for each robot. Prompts, task definitions, and evaluation metrics follow the protocol in Sec.~\ref{sec:composite}; additional training and implementation details are provided in the Appendix~\ref{app:vla_setup}.

As shown in Fig.~\ref{fig:pipeline_ablation}(b), PLanAR outperforms the fine-tuned VLA baseline in our evaluated sorting and stacking tasks. Although the VLA can learn useful visuomotor behaviors from task-specific demonstrations, it struggles with fine-grained semantic constraints and precise object placement. For example, in sorting, the policy may grasp a food item when instructed to pick a toy; in stacking, failures often result from imprecise grasps, unstable placements, or the lack of recovery after intermediate errors. In contrast, PLanAR explicitly decomposes the task, tracks symbolic state, and verifies execution outcomes, enabling more reliable recovery during long-horizon manipulation.

Fig.~\ref{fig:pipeline_ablation}(c) further compares PLanAR with TiPToP on the Franka Panda using the same disturbed fruit-sorting setup as Sec.~\ref{sec:pddl_verification_baseline}, as well as the block stacking task. In fruit sorting, TiPToP struggles under disturbance: after the second fruit is removed from the gripper and returned to the table, it does not recover from the resulting state mismatch. Additional failures also arise from incorrect grasps produced by Contact-GraspNet~\cite{sundermeyer2021contact}. On block stacking, TiPToP is unable to complete the task in our setup because cuTAMP~\cite{shen2025cutamp} does not generate feasible motion paths for the required stacking actions. In contrast, PLanAR achieves higher scores on both tasks by explicitly verifying execution outcomes, updating the task state, and replanning after disturbances or failed grasps. These results suggest that closed-loop reasoning and verification are particularly useful for tasks involving spatial constraints, sequential dependencies, and execution uncertainty. This robustness, however, comes at the cost of longer runtime, since PLanAR invokes additional verification and replanning steps after detected failures. Additional details are shown in the Appendix~\ref{app:tiptop}.

\section{Conclusion \& Limitations}

In this work, we presented PLanAR, a planning-language-grounded robot agent framework for real-world manipulation. PLanAR integrates VLM-based task understanding, PDDL-based planning, action-level verification, and closed-loop replanning into a model-agnostic pipeline. PLanAR provides a unified evaluation and deployment interface that plugs VLMs into different modules and exposes their strengths and weaknesses under a shared real-robot protocol. Results across lab, indoor, and in-the-wild scenes show that planning-language grounding structures action generation, while online verification detects failures, corrects state mismatches, and supports recovery. The same reasoning and verification logic transfers across robot embodiments by adapting only the low-level primitive interface. These findings suggest that robust robot agents should optimize for closed-loop consistency over richer open-loop reasoning.

Despite these promises, several limitations remain. First, PLanAR currently relies on predefined PDDL domains. Building a reliable domain requires iterative effort to refine predicates, action schemas, and effects. Second, verification remains latency-sensitive: dense VLM-based checking improves recovery, but replanning and grounding may require frequent model calls during execution. Future systems could reduce this overhead through asynchronous planning and grounding. Finally, PLanAR uses a compact set of robot primitives. Although these primitives support a diverse set of long-horizon tasks, they limit the range of skills that can be evaluated. Extending the primitive interface to more complex and freeform skills, including manipulation behaviors that cannot be reduced to grasp-and-place trajectories, is an important direction for scaling PLanAR.

\newpage

\bibliography{references}
\appendix
\newpage
\section{Implementation Details}
\label{app:Implementation_details}
\subsection{Implementation Overview}
\label{app:PLanAR_algo}
We describe the PLanAR pipeline in Algorithm~\ref{algo_1}. The system supports multiple action types, such as pick and place. We illustrate the main loop with pick; the remaining actions follow the same structure with minor, action-specific changes. For example, place performs only Detect and does not invoke AnyGrasp.

\begin{algorithm}[ht]
\caption{PLanAR Pipeline}
\label{algo_1}
\begin{algorithmic}[1]
\Statex \textbf{Require:}
\Statex \hspace{\algorithmicindent} Vision-Language Model $\mathcal{M}$
\Statex \hspace{\algorithmicindent} Task Instruction $l$
\Statex \hspace{\algorithmicindent} PDDL Domain File $D$
\Statex \hspace{\algorithmicindent} Text Prompt for Task Parser $p_{\text{task}}$
\Statex \hspace{\algorithmicindent} Text Prompt for Action Checker $p_\text{{checker}}$
\Statex \hspace{\algorithmicindent} 
Text Prompt for Grasp Planner $p_\text{{grasp}}$
\Statex \hspace{\algorithmicindent} 
Text Prompt for Goal Checker $p_\text{{goal}}$
\Statex \hspace{\algorithmicindent} Initial Observation $s_0^\text{sh}$ \algcmt{ ``sh" denotes shoulder camera.}
\Statex
\Statex \textcolor{blue}{\footnotesize{//} Query $\mathcal{M}$ to parse the task into $N$ actions $\{a_k\}_{k=1}^{N}$.}
\Statex \textcolor{blue}{\footnotesize{//} Each action specifies preconditions, target object, and effects.}

\State $\{a_k\}_{k=1}^{N}=\mathcal{M}(p_{\text{task}},\, l,\, D,\, s_0)$

\For{action $\{a_k\}_{k=1}^{N}$}
    \State Get observation $s_k^\text{sh}$
    \Statex \hspace{\algorithmicindent}  \textcolor{blue}{\footnotesize{//} Verify precondition.}    
    \If{\textbf{not}~$\mathcal{M}(p_{\text{checker}}, a_{k}^\text{pre}, s_k^\text{sh})$}
        \State Go to step~1
    \EndIf

    \Statex \hspace{\algorithmicindent}  \textcolor{blue}{\footnotesize{//} Detect the target object.}
    \State $\text{bbox}_k^{\text{sh}} = \text{Detect}(s_k^{\text{sh}}, a_{k}^\text{target})$
    \Statex \hspace{\algorithmicindent}  \textcolor{blue}{\footnotesize{//} Generate the grasp pose.}
    \State $\pi_k = \text{AnyGrasp}(s_k^\text{sh}, \text{bbox}_k^\text{sh}, a_k)$
    \Statex \hspace{\algorithmicindent}  \textcolor{blue}{\footnotesize{//} Evaluate the generated grasp.}
    \If{\textbf{not}~$\mathcal{M}(p_\text{grasp}, \pi_k, s_k^\text{sh})$}
        \State Move end-effector to $\pi_k$ to get wrist camera view
        \State Get observation $s_k^\text{wrist}$ from wrist camera
        \State $\text{bbox}_k^{\text{wrist}} = \text{Detect}(s_k^\text{wrist}, a_{k}^\text{target})$
            \State $\pi_k = \text{AnyGrasp}(s_k^\text{wrist}, \text{bbox}_k^\text{wrist}, a_k)$
    \EndIf
    \State Execute $\pi_k$
    \State Get post-action observation $s_k^{\text{post}}$
    \Statex \hspace{\algorithmicindent}  \textcolor{blue}{\footnotesize{//} Verify effect.}
    \If{\textbf{not}~$ \mathcal{M}(p_\text{checker}, a_{k}^\text{effect}, s_k^{\text{post}})$}
        \State Go to step~3
    \EndIf
    
\EndFor

\Statex  \textcolor{blue}{\footnotesize{//}~Verify task completion.}
    \If{\textbf{not}~$ \mathcal{M}(p_\text{goal},  s_N^{\text{post}})$}
        \State Go to step~1
    \EndIf
\end{algorithmic}
\end{algorithm}

\subsection{Qualitative Walkthrough and I/O Specifications}
\label{app:walk_through}

We utilize a unified interface for all VLM components. To ensure reproducibility, we provide a concrete walkthrough of the main modules.

\begin{figure*}[!h] 
    \centering
    \includegraphics[width=0.9\linewidth]{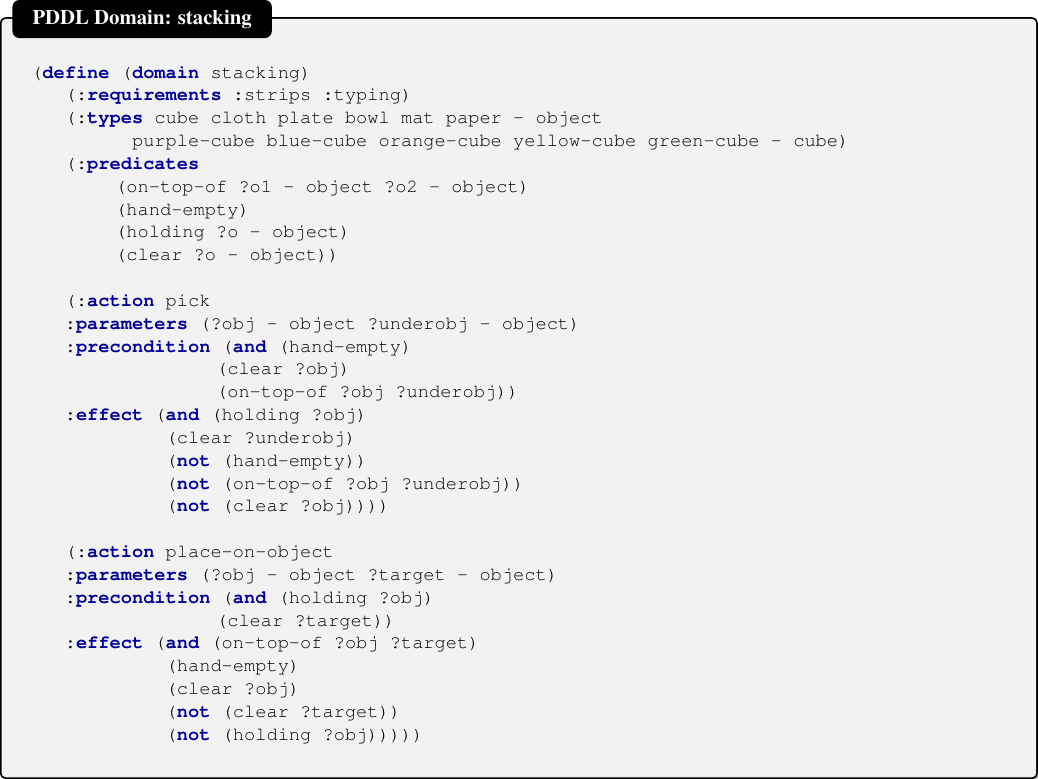} 
    \caption{The defined PDDL domain of stacking task.}
    \label{fig:pddl_domain}
\end{figure*}

\begin{figure}[ht]
    \centering
    \includegraphics[width=0.9\linewidth]{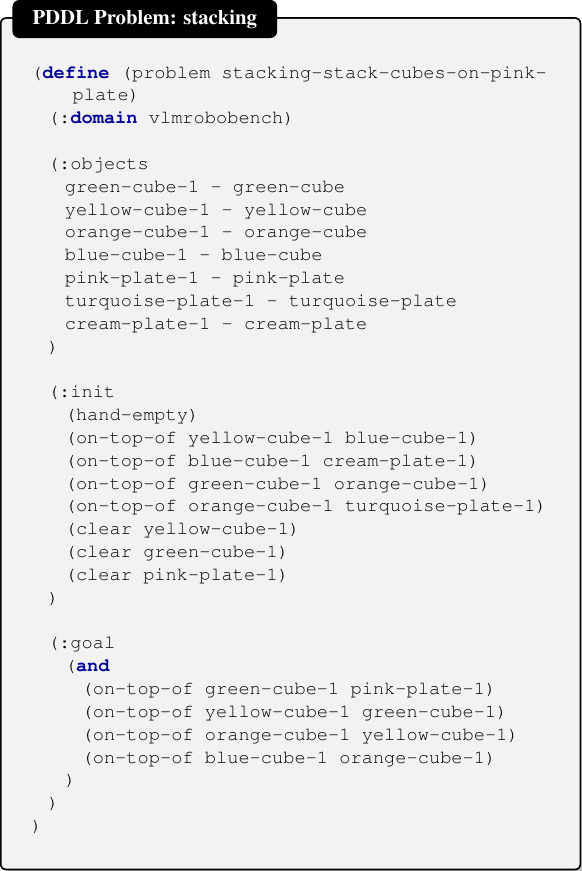} 
    \caption{The PDDL Problem File Generated by Gemini 3 Flash. Stack the cubes on the pink plate from bottom to top: Orange Blue Yellow and Green cubes}
    \label{fig:pddl_problem}
\end{figure}

\begin{figure}[ht] 
    \centering
    \includegraphics[width=0.9\linewidth]{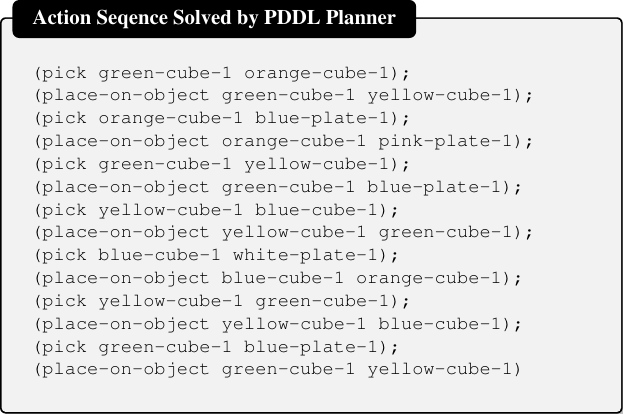} 
    \caption{Action sequence obtained by solving the updated domain file with the generated problem file.}
    \label{fig:action_sequence}
\end{figure}

% 1. Task Parser Logic
\textbf{Task Parser.} To bridge natural-language instructions and symbolic planning, the task parser facilitates the translation from instruction to PDDL. As illustrated in Table~\ref{tab:task_parser_prompt}, it operates as follows:
\begin{enumerate}[label=(\roman*)]
    \item {Input:} The model receives the domain definition (Fig.~\ref{fig:pddl_domain}), scene observation, and a user instruction (e.g., ``stack the cubes...").
    \item {Reasoning:} The VLM generates a PDDL problem file (Fig.~\ref{fig:pddl_problem}) that defines the initial state and goal conditions based on the visual observation.
    \item {Output:} A classical planner, Fast Downward~\cite{helmert2006fast}, solves this PDDL to produce an executable action sequence (Fig.~\ref{fig:action_sequence}).
\end{enumerate}

\begin{table*}[t]
\centering
\begin{tabular}{|p{0.95\linewidth}|}
\hline
\textbf{The input request contains:} \\
\begin{itemize}[leftmargin=*, nosep]
    \item A string describing the multi-stage manipulation task.
    \item An image of the current table-top environment captured from a shoulder camera.
    \item A PDDL domain file defining available predicates and actions.
    \item An example problem in both natural language and PDDL format for few-shot learning.
\end{itemize} \\
\textbf{The output response is a JSON object containing:} \\
\begin{itemize}[leftmargin=*, nosep]
    \item \textbf{objects\_identified}: A list of objects identified from the scene. Each object has a \texttt{name} (PDDL identifier, e.g., ``blue-block-1''), \texttt{type} (object category), and \texttt{description} (natural language description of the object).
    \item \textbf{reasoning}: A string providing brief analysis of task-relevant objects, their types, spatial relationships, and task requirements.
    \item \textbf{updated\_domain}: The complete PDDL domain file with new object types added if the task mentions types not in the original domain.
    \item \textbf{problem\_pddl}: A complete PDDL problem file containing the \texttt{:objects} section listing all identified objects, \texttt{:init} section describing the current state, and \texttt{:goal} section specifying the desired final state.
\end{itemize} \\
\hline
\end{tabular}
\caption{The high-level reasoning prompt for task parser. It analyzes the scene image and task instruction to generate a structured PDDL representation that can be solved by a classical planner.}
\label{tab:task_parser_prompt}
\end{table*}

% 2. Action Checker Logic
\textbf{Action Checker.} To ensure closed-loop reasoning and robustness, the action checker validates both preconditions and effects. Table~\ref{tab:action_checker_prompt} demonstrates this capability.

\begin{figure*}[t]
    \centering

    \begin{subfigure}[t]{0.32\linewidth}
        \centering
        \includegraphics[width=\linewidth]{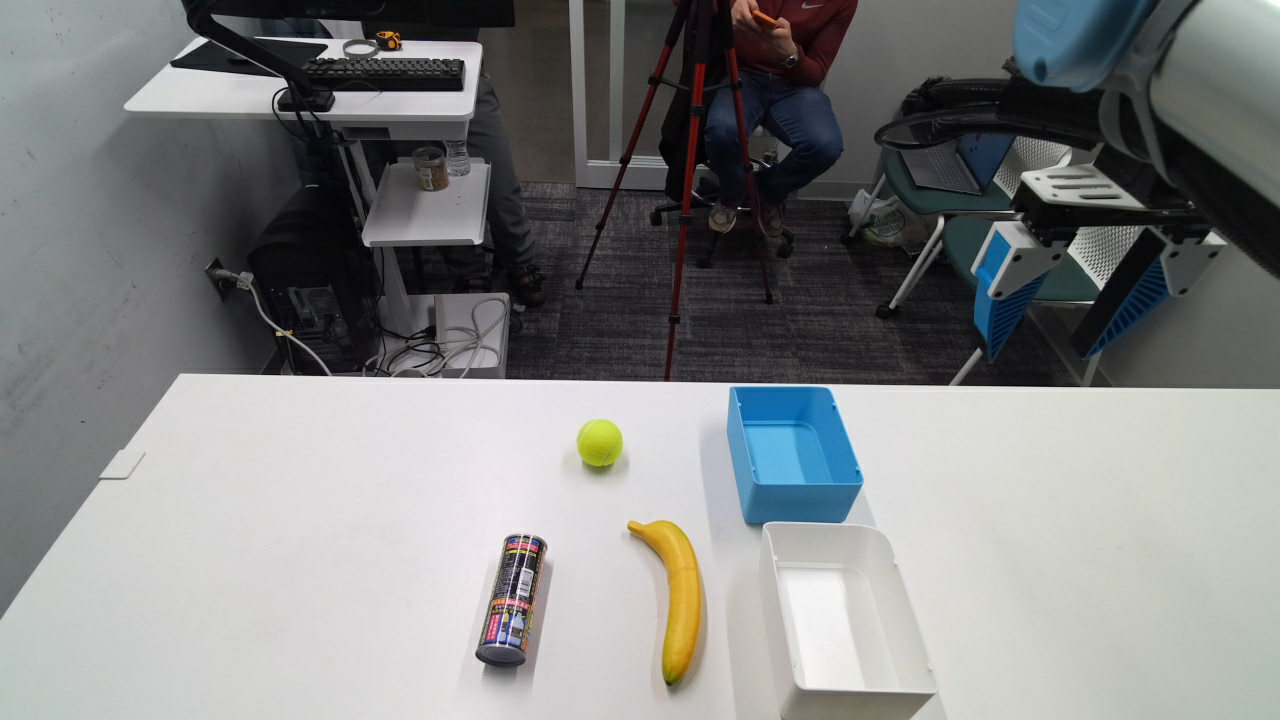}
        \caption{Precondition check (pick can-1)}
        \label{fig:precond_image_can}
    \end{subfigure}
    \hfill
    \begin{subfigure}[t]{0.32\linewidth}
        \centering
        \includegraphics[width=\linewidth]{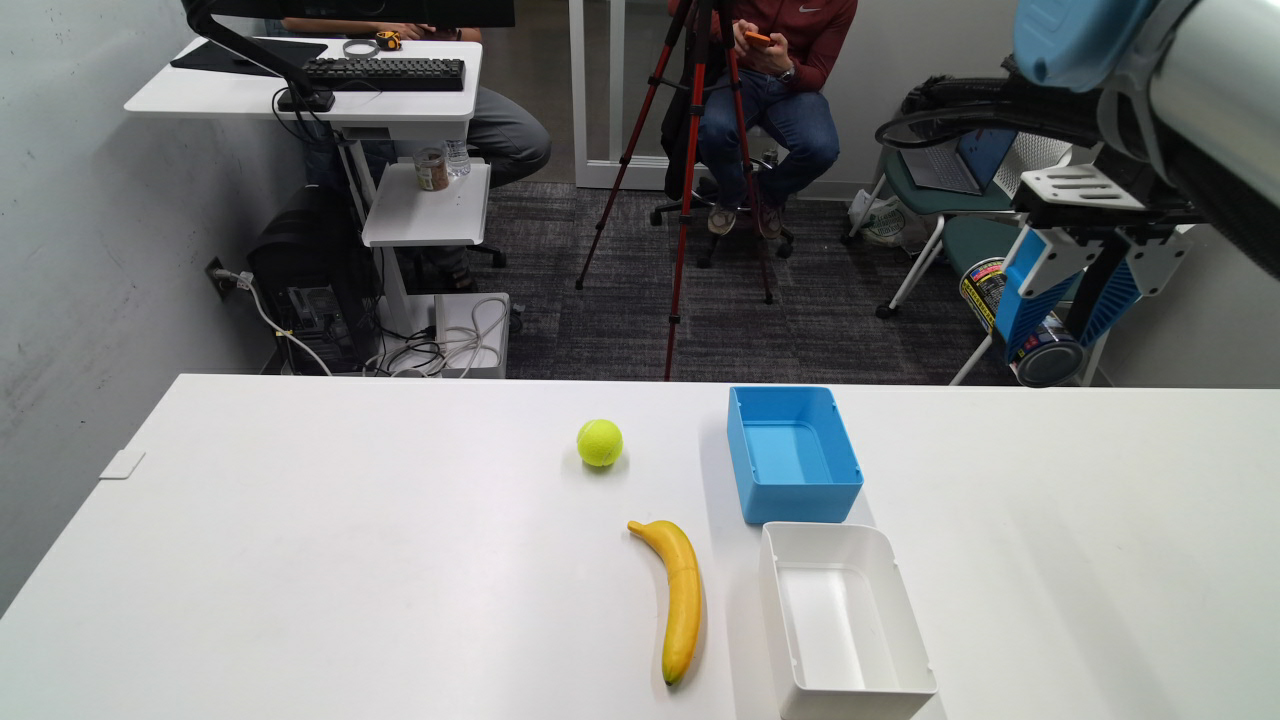}
        \caption{Effect check (pick can-1)}
        \label{fig:effect_image_can}
    \end{subfigure}
    \hfill
    \begin{subfigure}[t]{0.32\linewidth}
        \centering
        \includegraphics[width=\linewidth]{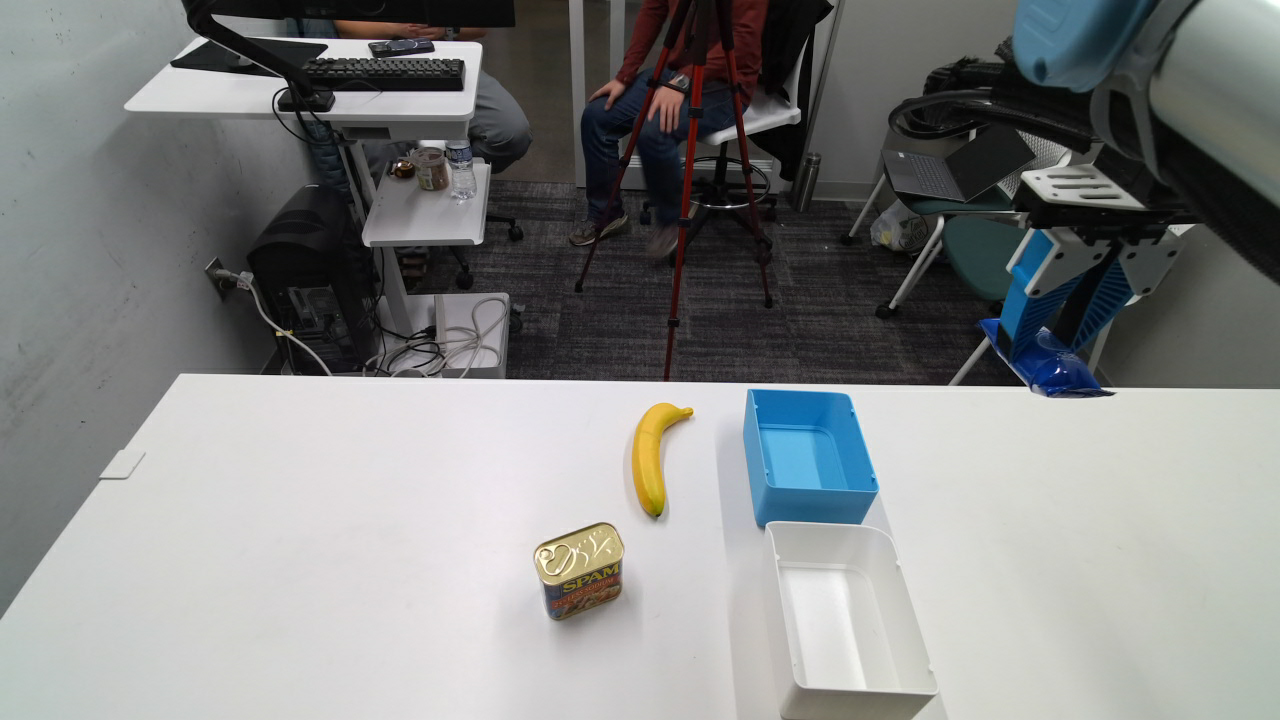}
        \caption{Effect check (pick oreo-pack-1)}
        \label{fig:effect_image_oreo}
    \end{subfigure}

    \caption{Precondition/effect checks for pick actions across objects.}
    \label{fig:action_checker_input_example}
\end{figure*}

\begin{figure*}[t]
    \centering
    \includegraphics[width=0.9\linewidth]{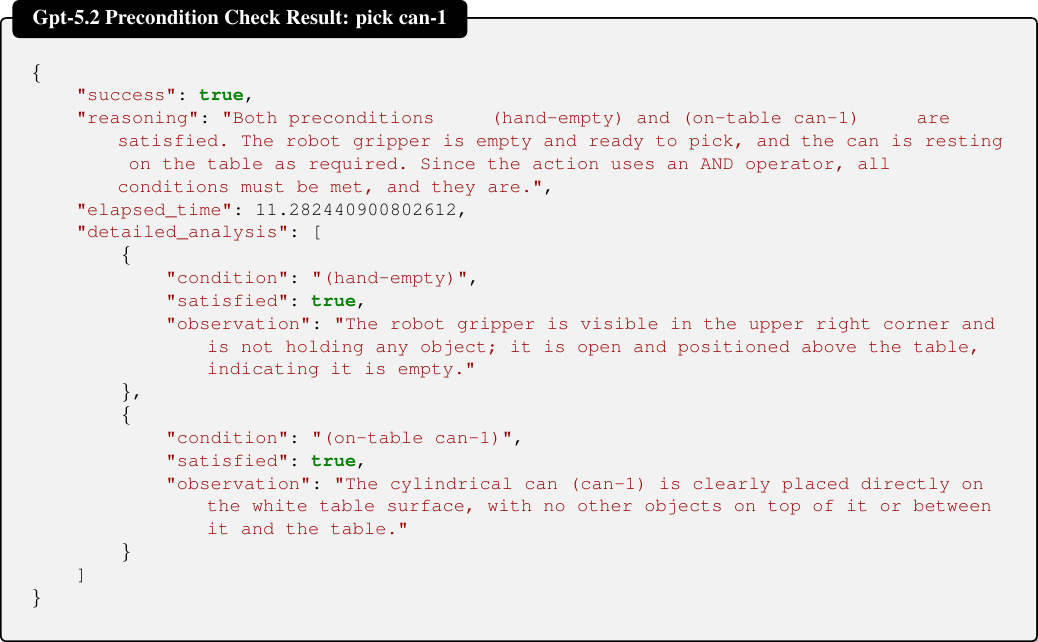}
    \caption{Precondition check for picking can (success, GPT-5.2).}
    \label{fig:ac_pre_can_gpt}
\end{figure*}

\begin{figure*}[t]
    \centering
    \includegraphics[width=0.9\linewidth]{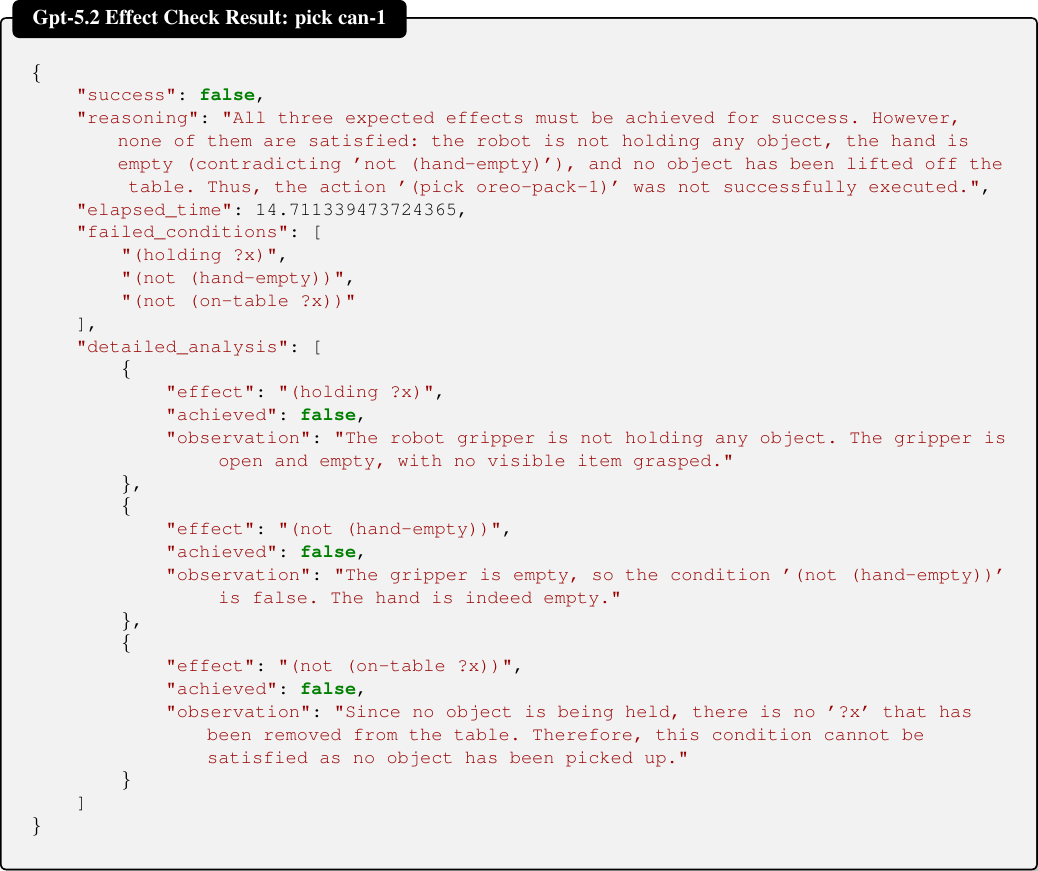}
    \caption{Effect check for picking can (fail, GPT-5.2).}
    \label{fig:ac_eff_can_gpt}
\end{figure*}

\begin{figure*}[t]
    \centering
    \includegraphics[width=0.9\linewidth]{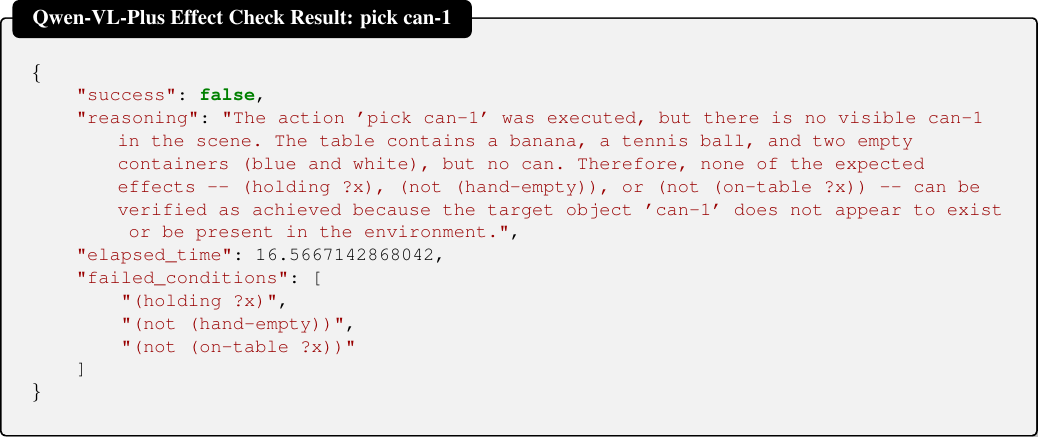}
    \caption{Effect check for picking can (fail, Qwen-VL-Plus).}
    \label{fig:ac_eff_can_qwen}
\end{figure*}

\begin{figure*}[t]
    \centering
    \includegraphics[width=0.9\linewidth]{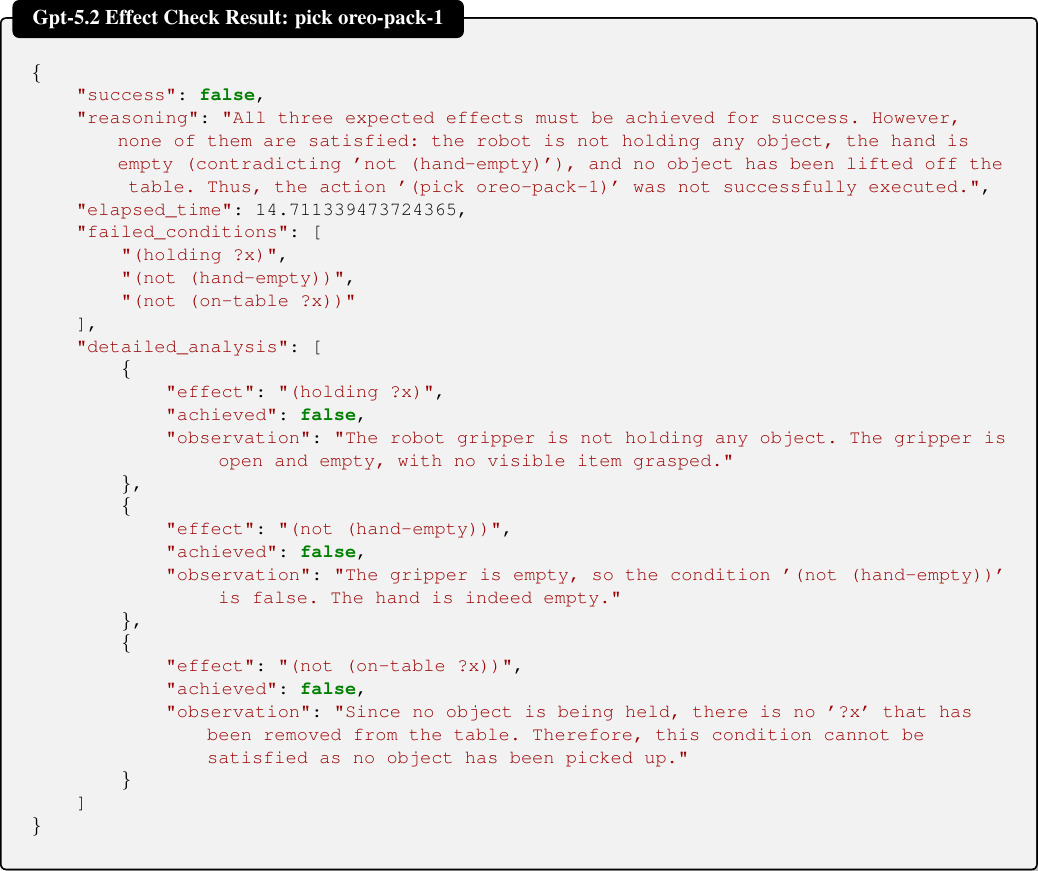}
    \caption{Effect check for picking oreo pack (fail, GPT-5.2).}
    \label{fig:ac_eff_oreo_gpt}
\end{figure*}

A detailed input example for the action checker is shown in Fig.~\ref{fig:action_checker_input_example}, including representative precondition and effect queries for pick actions under the shoulder camera view. Figs.~\ref{fig:ac_pre_can_gpt}-\ref{fig:ac_eff_oreo_gpt} further visualize corresponding JSON outputs from different VLMs, illustrating how the checker reports per-predicate satisfaction and diagnoses failure cases when evaluated effects are not correct.

% action checker prompt
\begin{table*}[t]
\centering
\begin{tabular}{|p{0.95\linewidth}|}
\hline
\textbf{The input request contains:} \\
\begin{itemize}[leftmargin=*, nosep]
    \item A PDDL action to be executed (e.g., \texttt{(pick apple-1)}).
    \item The preconditions for the action in PDDL format (e.g., \texttt{(and (hand-empty) (clear apple-1))}).
    \item An image of the current scene captured from a shoulder camera showing the robot gripper and workspace.
    \item A reference guide explaining PDDL predicates such as \texttt{hand-empty}, \texttt{holding}, \texttt{clear}, \texttt{on-table}, and \texttt{on-top-of}.
\end{itemize} \\
\textbf{The output response is a JSON object containing:} \\
\begin{itemize}[leftmargin=*, nosep]
    \item \textbf{conditions\_analysis}: A list of individual condition evaluations. Each entry contains a \texttt{condition} (the exact predicate string), \texttt{satisfied} (boolean indicating if the condition holds), and \texttt{observation} (visual evidence from the image supporting the evaluation).
    \item \textbf{success}: A boolean indicating whether ALL preconditions are satisfied based on logical operators (and/or/not).
    \item \textbf{reasoning}: A string providing the overall conclusion explaining why the preconditions are or are not satisfied.
    \item \textbf{failed\_conditions}: A list of predicate strings that are NOT satisfied (empty if all conditions are met).
\end{itemize} \\
\hline
\end{tabular}
\caption{The precondition verification prompt for action checker. It visually validates whether the current scene satisfies all preconditions before executing a planned action, enabling closed-loop error detection.}
\label{tab:action_checker_prompt}
\end{table*}

% action checker input images

% ---- Action checker qualitative examples  ----

\begin{figure*}[ht] 
    \centering
    \includegraphics[width=0.9\linewidth]{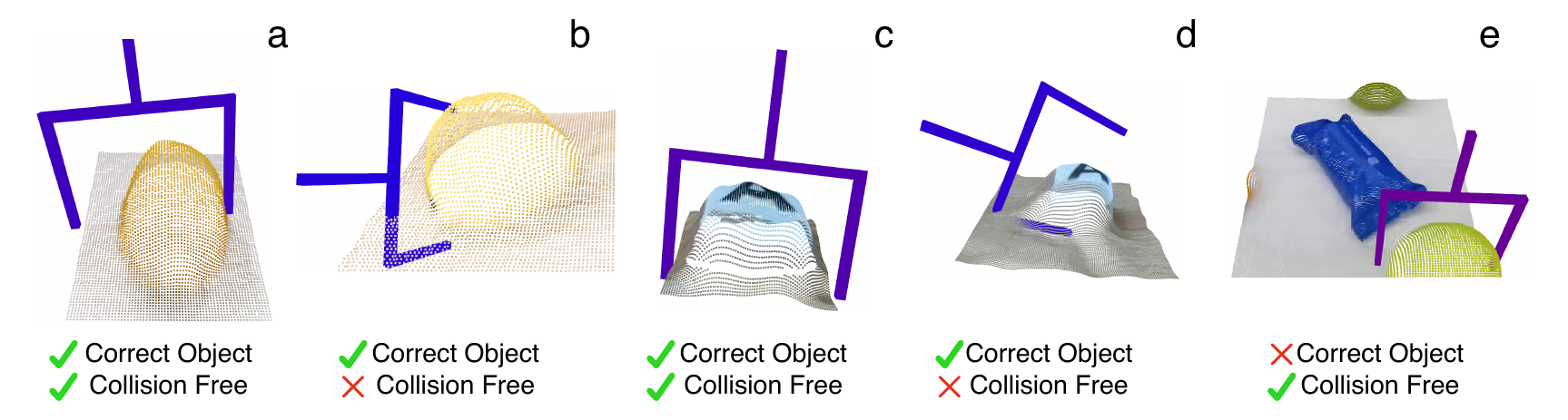} 
    \caption{Evaluation of proposed grasp poses. Ground-truth labels are provided below each example.}
    \label{fig:grasp_planner_input}
\end{figure*}

\begin{figure*}[ht] 
    \centering
    \includegraphics[width=0.9\linewidth]{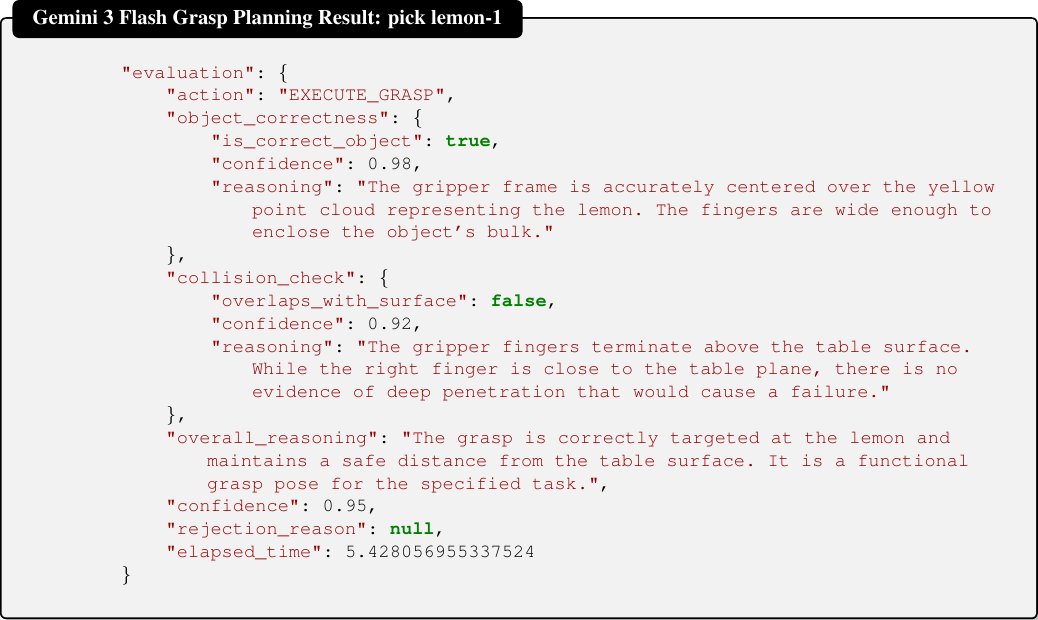} 
    \caption{Gemini 3 Flash output for example a in Fig.~\ref{fig:grasp_planner_input},  planning the pick lemon-1 action.}
    \label{fig:grasp_planner_1}
\end{figure*}

\begin{figure*}[ht] 
    \centering
    \includegraphics[width=0.9\linewidth]{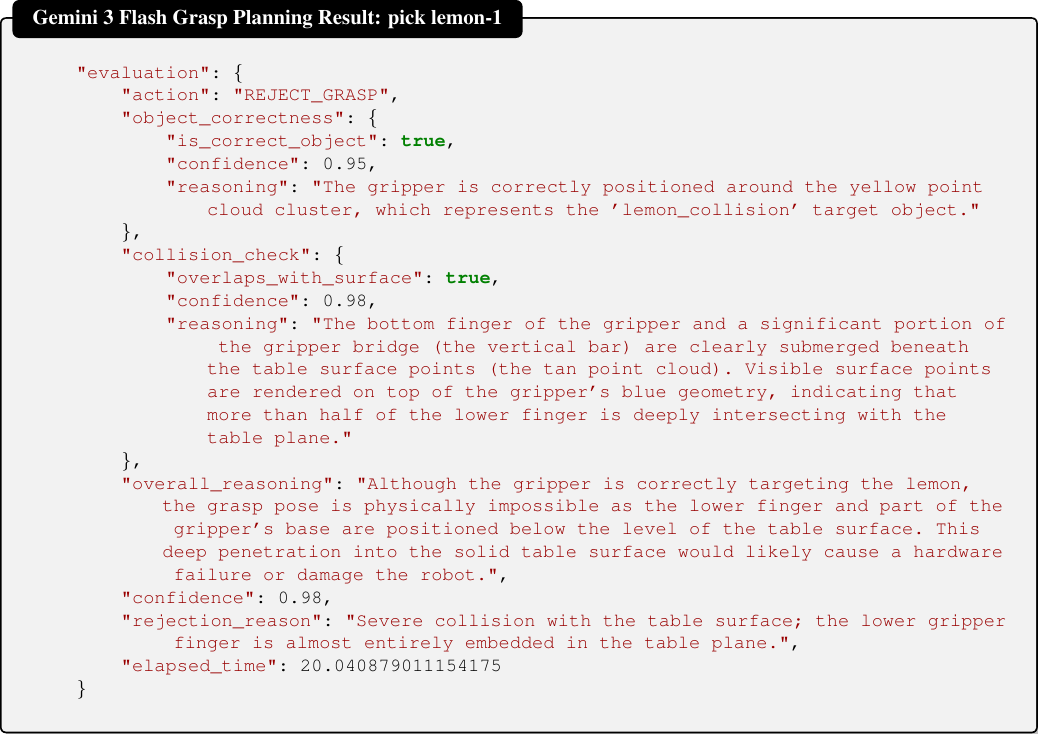} 
    \caption{Gemini 3 Flash output for example b in Fig.~\ref{fig:grasp_planner_input},  planning the pick lemon-1 action.}
    \label{fig:grasp_planner_2}
\end{figure*}

\begin{figure*}[ht] 
    \centering
    \includegraphics[width=0.9\linewidth]{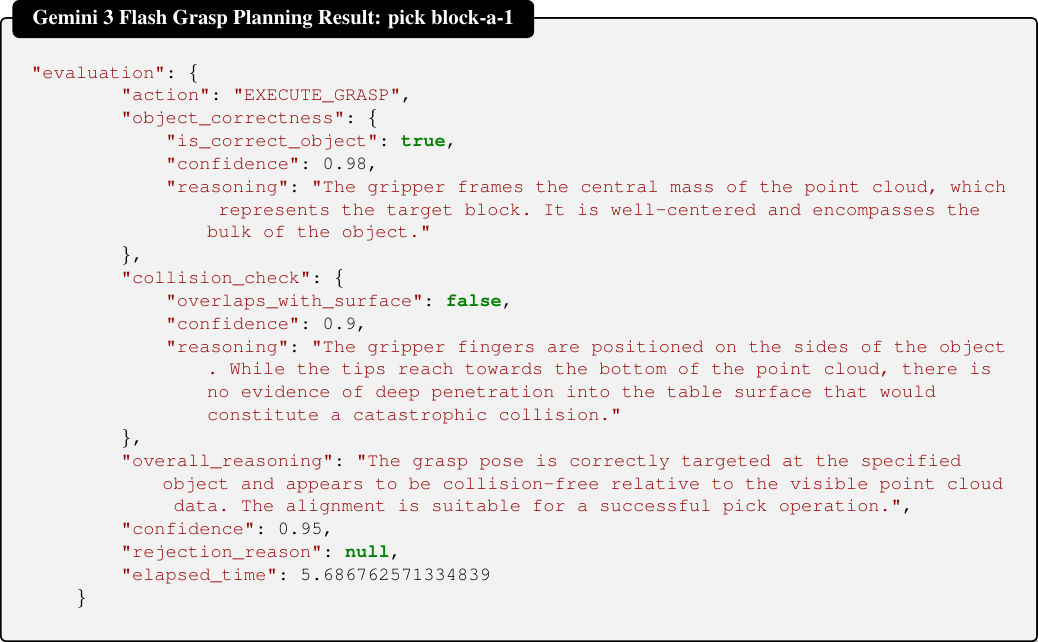} 
    \caption{Gemini 3 Flash output for example c in Fig.~\ref{fig:grasp_planner_input},  planning the pick block-a-1 action.}
    \label{fig:grasp_planner_3}
\end{figure*}

\begin{figure*}[ht] 
    \centering
    \includegraphics[width=0.9\linewidth]{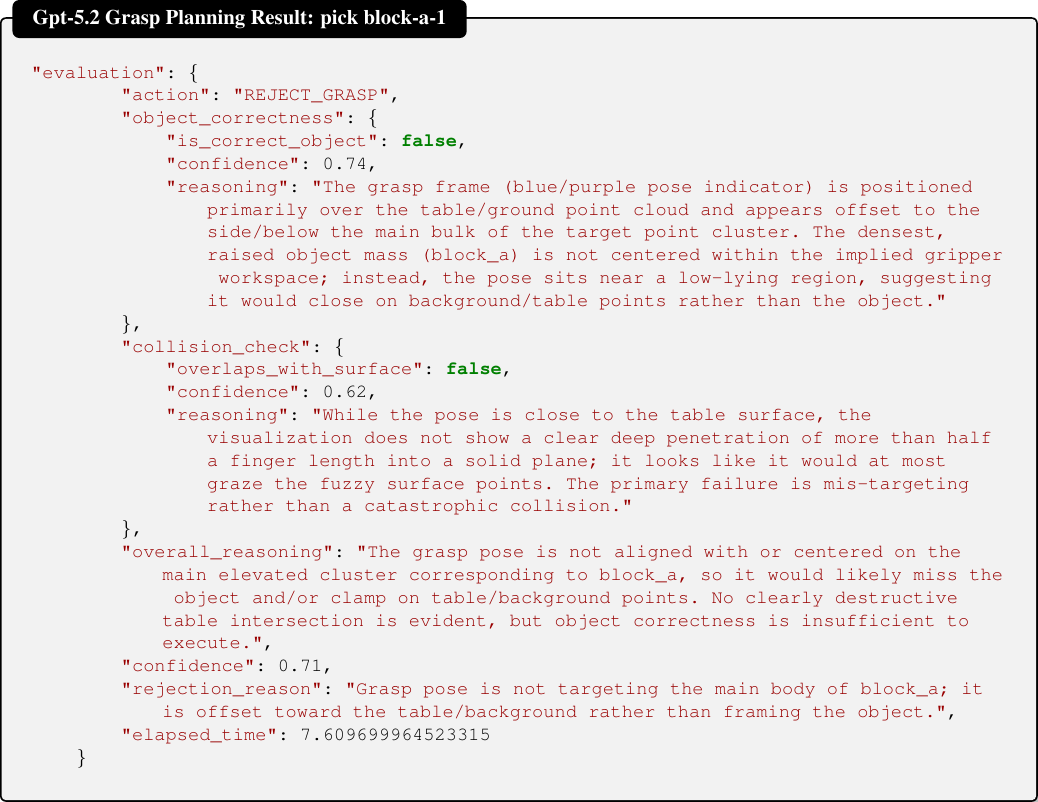} 
    \caption{GPT-5.2 output for example d in Fig.~\ref{fig:grasp_planner_input},  planning the pick block-a-1 action.}
    \label{fig:grasp_planner_4}
\end{figure*}

\begin{figure*}[ht] 
    \centering
    \includegraphics[width=0.9\linewidth]{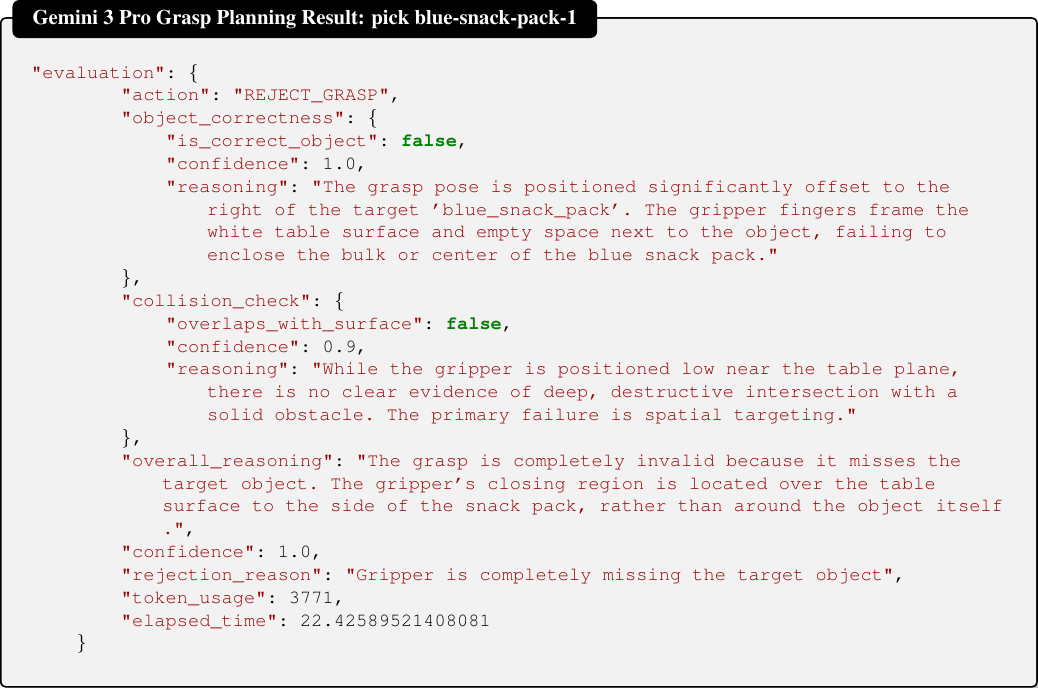} 
    \caption{Gemini 3 Pro output for example e in Fig.~\ref{fig:grasp_planner_input},  planning the pick blue-snack-pack-1 action.}
    \label{fig:grasp_planner_5}
\end{figure*}

\textbf{Grasp Planner.} Fig.~\ref{fig:grasp_planner_input} illustrates the input used to evaluate semantic correctness and physical feasibility: the proposed grasp pose visualized in the local point cloud. We provide ground-truth executability labels (a-e) for better understanding. The corresponding model reasoning and judgments are detailed in Figs.~\ref{fig:grasp_planner_1}-\ref{fig:grasp_planner_5}. The model evaluates whether the pose targets the correct semantic object and is collision-free. Crucially, a \texttt{REJECT} decision triggers the system to switch to the wrist camera for a close-up re-evaluation and re-planning.


\section{Experiment Details}
\label{app:exp_details}

\begin{figure}[!h]
  \centering
  % Replace with your actual filename
  \includegraphics[width=1\textwidth, trim=5 60 3 0, clip]{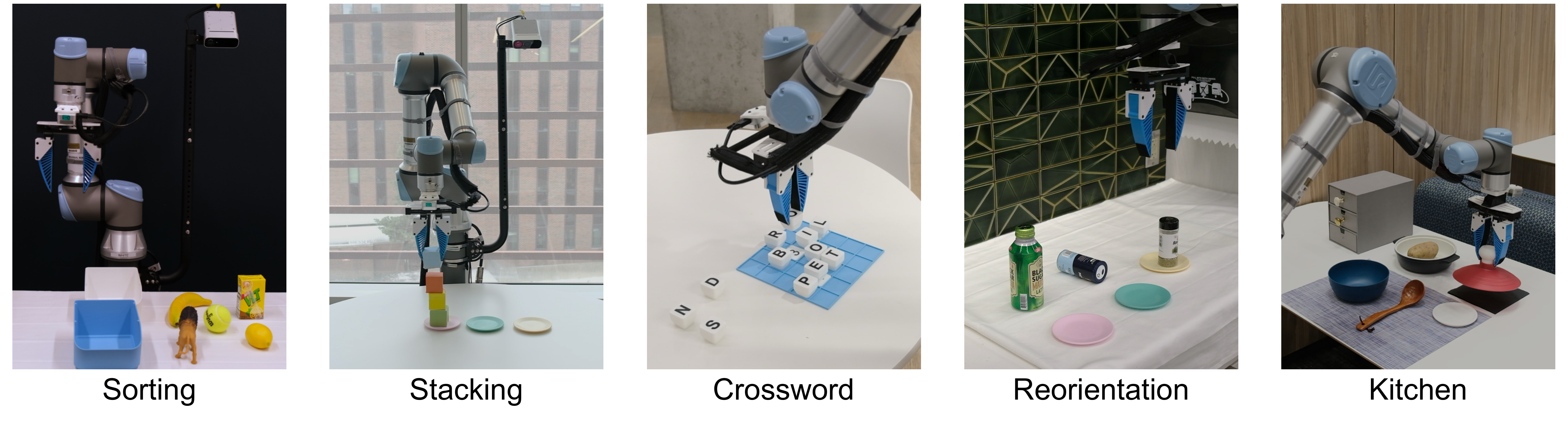}
  \caption{Manipulation tasks used for real-world evaluation. These tasks are selected to cover a range of challenges, including spatial reasoning, vision-language grounding, and long-horizon planning.}
  \label{fig:task}
  \vspace{-1em} % Optional: Adjusts space below figure to save paper space
\end{figure}

\subsection{Task Progress Score}
\label{app:task_setup_metrics}

In both the compositional vs. single-VLM experiments and the ablation study, we measure performance using a task progress score derived from a structured action-sequence rubric. The score measures how much of the intended plan was completed, while penalizing unnecessary actions and final goal-check failures, making it more informative than binary success. This structured scoring system allows for consistent evaluation and comparison of task performance.

Concretely, let $N$ be the total number of executed actions, $N_{\text{done}}$ the number of actions that match the required action sequence, and $N_{\text{extra}}$ the number of redundant actions. Denote $p \in \{0,1\}$ as a binary penalty for goal-checker error, where $p = 1$ indicates that the goal checker failed to correctly assess the final task state. We define:
\begin{align*}
\text{Score} \;=\; \frac{N_{\text{done}} - N_{\text{extra}}}{N} \;-\; 0.1\, p \, .
\end{align*}

The redundant-action penalty is important because VLMs sometimes introduce redundant goals when generating the problem PDDL, which can trigger unnecessary steps, this behavior is especially common in the crossword task.

% =========================
% Module-level Evaluation
% =========================
\subsection{Module Evaluation}
We report runtime as an averaged estimate from randomized evaluations conducted across multiple time periods. While latency for cloud-hosted models is largely determined by the provider’s service, our goal is to provide an intuitive sense of query-time differences across models, enabling practitioners to make informed deployment choices for robotic agents based on their efficiency requirements.

\begin{table}[H]
\centering
\caption{Performance of single-VLM pipelines on the sorting task. Each pipeline uses one VLM for all modules and is evaluated over 20 trials across 6 object setups.}
\label{tab:single_vlm_end2end}
\begin{tabular}{lccc}
\toprule
\textbf{Model Pipeline} & \textbf{Success Rate} & \textbf{Time} & \textbf{Tokens} \\
\midrule
% --- Google Gemini Series ---
Gemini 3 Flash & \textbf{80.0 \%} & 334~s & $ $67.0k \\
Gemini 3 Pro & 10.0  \%& 1283~s & $ $139.0k \\
Gemini 3 Pro + LangSAM & 70.0  \%& 524~s & $ $80.9k \\
Gemini Robotics & 50.0  \%& 382~s & \textbf{40.6k} \\
\midrule
% --- Qwen Series ---
Qwen-VL-Max & 0.0  \%& -- & -- \\
Qwen3-VL-Plus  & 25.0  \%& 538~s & $ $43.0k \\
\midrule
% --- OpenAI Series ---
GPT-5.2 + LangSAM & 0.0  \%& -- & -- \\
GPT-4o + LangSAM & 5.0  \%& \textbf{299~s}& $ $49.6k \\
\bottomrule
\end{tabular}%
\end{table}

\subsection{Module Evaluation Details}
\label{app:module_eval}

\textbf{Data Annotation:}
We report the data annotation details used in our module-level evaluation.

1. Action Checker.
We collect action-checker evaluation data from real rollouts of the PLanAR pipeline, resulting in 61 verification instances across 7 different scenes and 5 tasks. For each model, we run three trials and report the average success rate, latency, and token usage.

2. Grasp Planner.
We curate 50 grasp-planning instances covering diverse objects. Executability is labeled from real-world outcomes, where a grasp is considered valid if it picks the correct object and executes without collisions. Each model is evaluated over three runs. Notably, local models frequently collapse to near-constant true/false outputs, indicating limited capability as grasp planners under this protocol.

3. Object Detector.
We collect 21 real-world camera inputs and annotate object regions using SAM masks \cite{kirillov2023segment} and corresponding bounding boxes. We compute (i) point-in-mask success rate, based on whether the VLM-predicted point lies inside the SAM mask, and (ii) bounding-box IoU, defined as the overlap area divided by the union area between the predicted box and the annotated box. Each model is tested three times and we report averaged results.

4. Goal Checker.
For task-level parsing evaluation, we record 16 final-scene cases and ask the VLM to judge whether the task goal is satisfied. Each model is evaluated with three runs. Ground-truth labels are provided by five human annotators, who determine success based on the goal specification and the task prompt.

\begin{figure*}[t]
    \centering
    \includegraphics[width=0.92\linewidth]{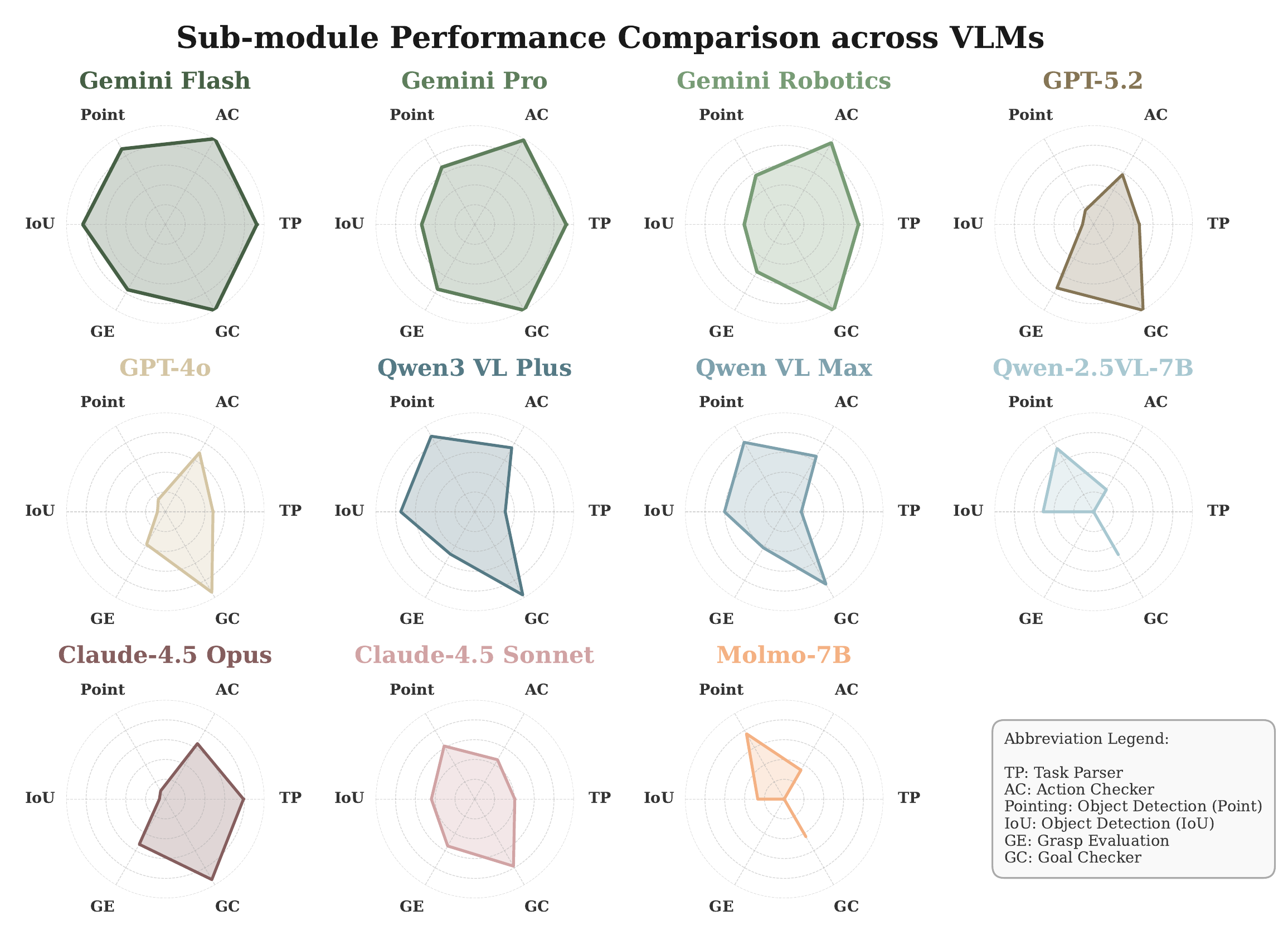}
    \caption{{Modular evaluation radar chart.} Normalized per-module performance across the pipeline (higher is better). Each axis reports the corresponding module score.}
    \label{fig:module_radar}
\end{figure*}

\begin{figure*}[t]
    \centering
    \includegraphics[width=0.95\linewidth]{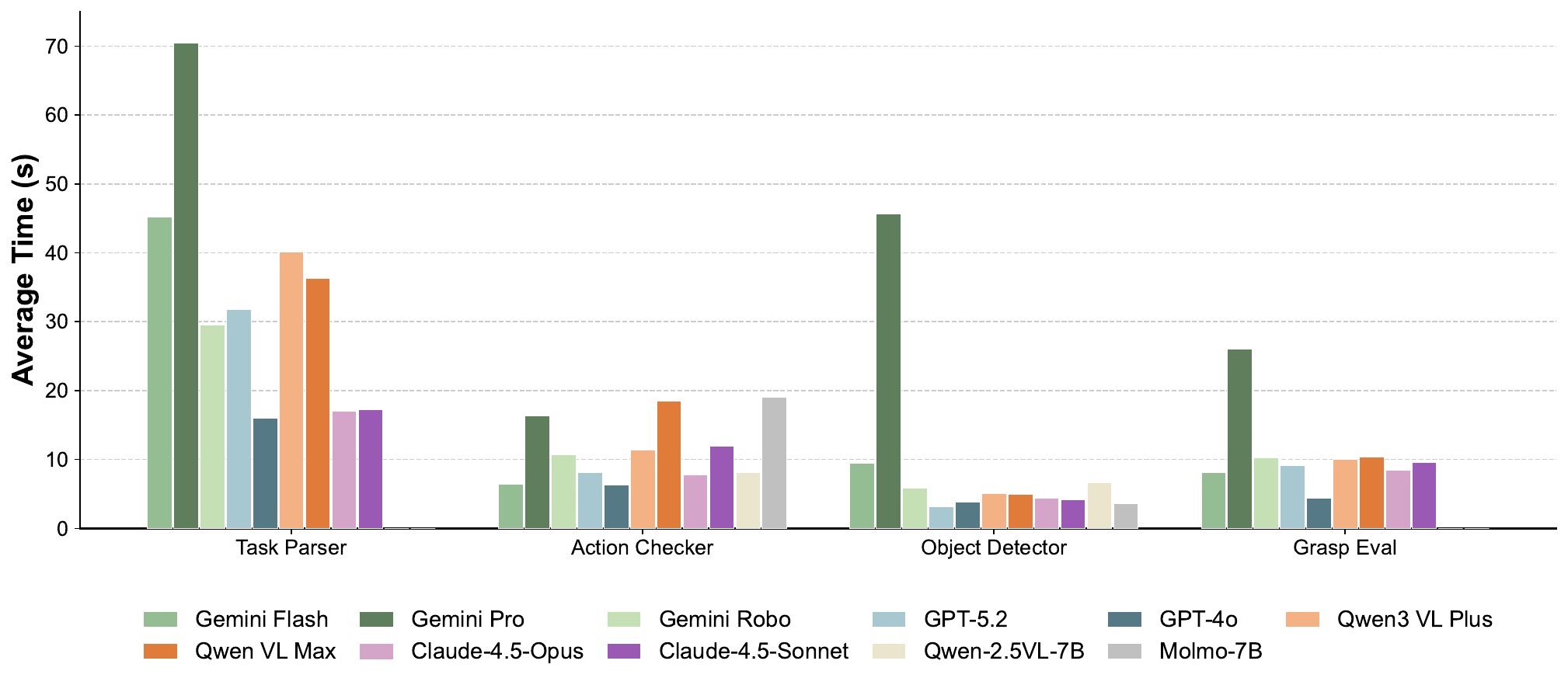}
    \caption{{Latency comparison across VLMs.} Average runtime per module evaluation query.}
    \label{fig:submodule_time}
\end{figure*}

\begin{figure*}[t]
    \centering
    \includegraphics[width=0.92\linewidth]{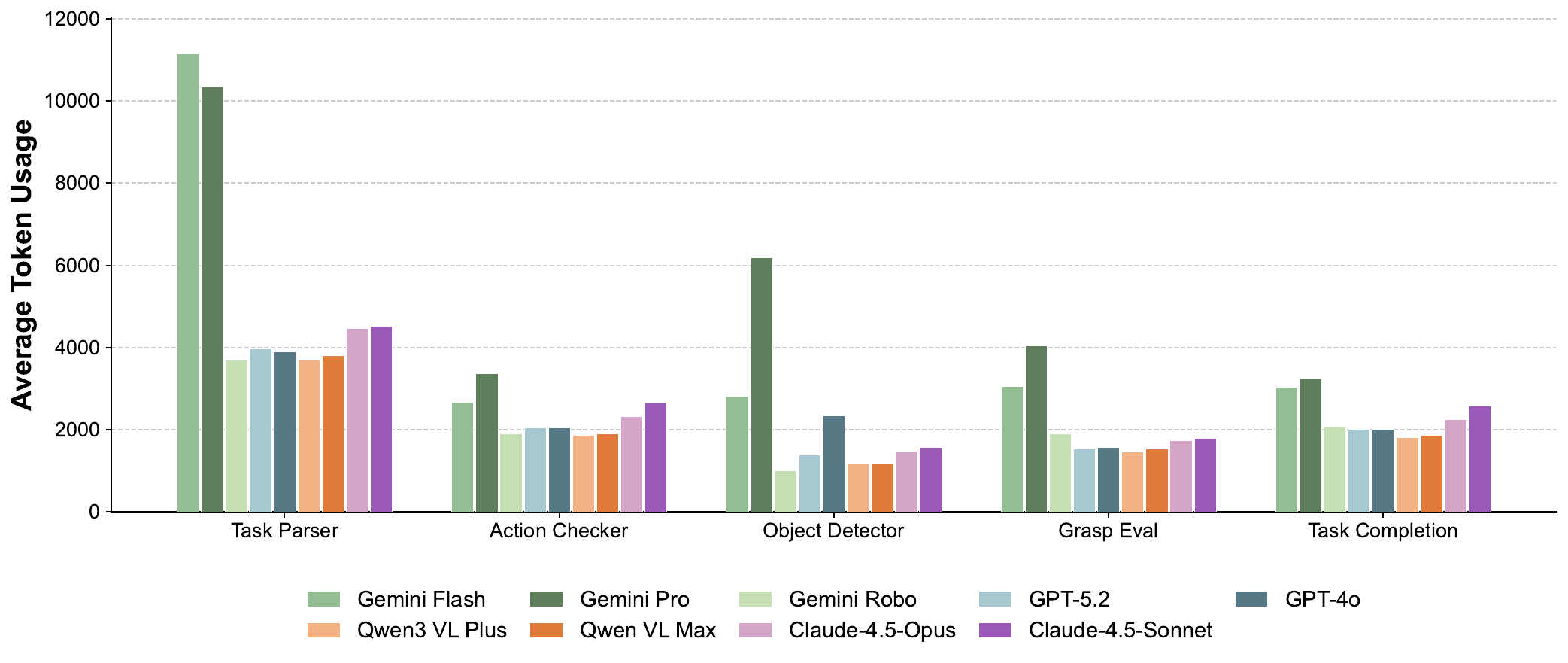}
    \caption{{Token usage comparison across VLMs.} Average token consumption per module evaluation query.}
    \label{fig:submodule_token}
\end{figure*}

\textbf{Module Evaluation Results:}
We summarize per-module performance with a radar chart in Fig.~\ref{fig:module_radar}. Each axis corresponds to a module-level metric, and we normalize all scores to $[0,1]$ so that larger values indicate better performance. We also report efficiency across VLM backends using the average latency in Fig.~\ref{fig:submodule_time} and token usage in Fig.~\ref{fig:submodule_token}, both aggregated over the same module evaluation runs.

\textbf{Human Evaluation Interface:}
Evaluating the task parser in open-world scenes benefits from a human study, since fully automatic evaluation can rely on brittle, hand-crafted heuristics. We therefore conduct a human study and build a lightweight labeling UI that supports efficient, reproducible evaluation while ensuring fair, blinded comparison across models.

\begin{figure*}[t]
    \centering
    % Replace with your screenshot
    \includegraphics[width=0.95\linewidth]{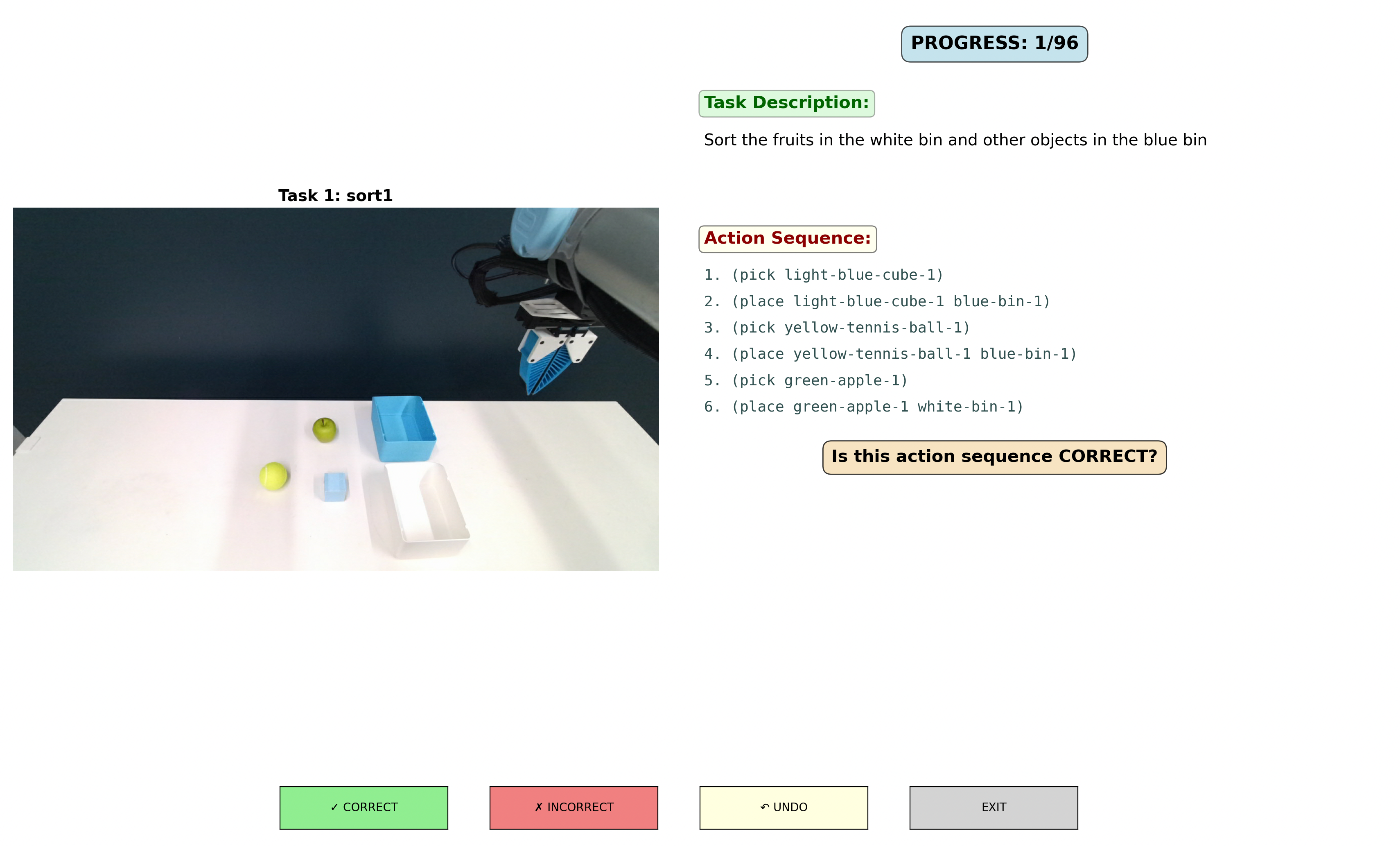}
    \caption{Human evaluation UI.}
    \label{fig:human_ui}
\end{figure*}

As shown in Fig.~\ref{fig:human_ui}, participants are shown only (i) the task goal, (ii) the initial scene image at the start of the episode, and (iii) the action sequence produced by the task parser. The model identity is hidden. Participants judge whether the provided action sequence can reasonably achieve the goal under the given initial scene conditions.

We evaluate four task categories: sorting, stacking, crossword, and kitchen. We recruited five participants and collected 468 action-sequence judgments in total. To improve reliability, we perform three full passes over the complete set and report results aggregated across rounds.

Table~\ref{tab:task_parser_task_list} lists the natural language instructions used in our task parser evaluation, covering the four evaluation categories described above.
% --- Task list table (single column, tabularx style) ---
\begin{table}[t]
\centering
\small
\renewcommand{\arraystretch}{1.5}
\begin{tabularx}{\linewidth}{l | X}
\hline
\textbf{Task} & \textbf{Language Description} \\ \hline

Sort & Sort the fruits into the white bin and all other objects into the blue bin. \\ \hline

Stack (3 cubes) & Stack the cubes on the pink plate from bottom to top: Green, Orange, and Blue. \\ \hline

Stack (4 cubes) & Stack the cubes on the pink plate from bottom to top: Green, Yellow, Orange, and Blue. \\ \hline

Crossword & Fill the numbered slots using the provided blocks to solve the crossword puzzle. You do not need to use all blocks or all slots. \\ \hline

Kitchen A & Put the spice bottle into the top drawer and close it. \\ \hline

Kitchen B & Place the blue snack pack in the top drawer, then move the spice bottle from the drawer to the table, and finally close the drawer. \\ \hline

Kitchen C & A chicken leg is in the pot. Take out the chicken leg, place it in the bowl, then put the spice bottle back into the top drawer and close the drawer. \\ \hline

\end{tabularx}
\caption{Task list and language descriptions.}
\label{tab:task_parser_task_list}
\end{table}

\subsection{Ablation Study: Action Checker and Grasp Planner}\label{app:closed_loop_ablation}

\begin{figure}[H]
  \centering
  \includegraphics[width=\linewidth]{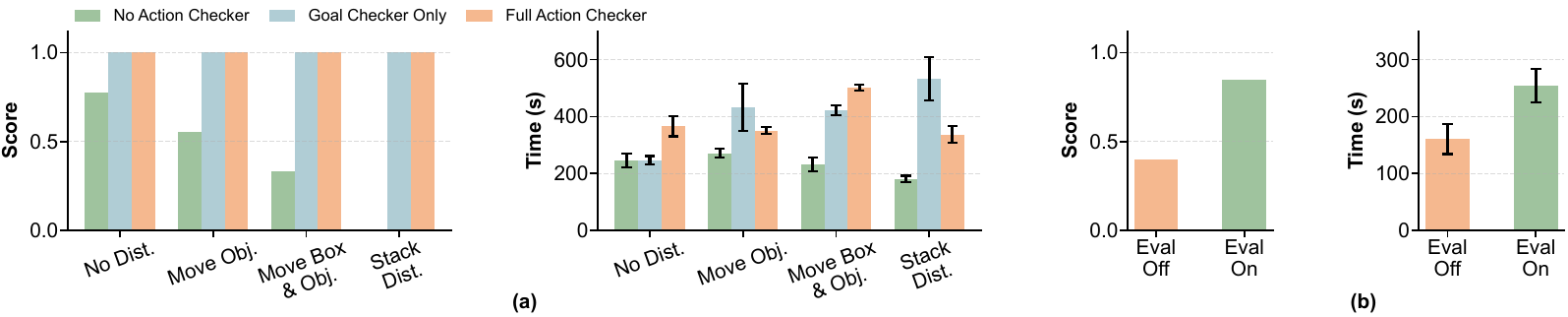}
  \caption{Ablation results for action checking and grasp planning. We report task progress score and execution time under different verification and grasp-planning settings. Error bars denote standard deviation.}
  \label{fig:ablation_all}
\end{figure}

In this experiment, we characterize the robustness–efficiency trade-off via targeted ablations of two components. First, we compare three action-checking configurations: no checker, goal checker only, and full action checker, under controlled disturbances (Fig.~\ref{fig:ablation_all}(a)). Second, we ablate the grasp planner by enabling or disabling grasp evaluation, as shown in Fig.~\ref{fig:ablation_all}(b).

We run the action checker ablations on two tasks: sorting, where actions are weakly coupled, and sequential stacking, where actions are strongly dependent and early failures affect all subsequent actions. For the grasp planner, we ablate grasp evaluation on sorting in clustered scenes, where imperfect bounding boxes often include multiple objects and collision-prone grasps are common.

Two patterns emerge. First, grasp evaluation is critical in cluttered scenes. When it is disabled, task success drops sharply because the system is more likely to execute collision-prone grasps or pick the wrong object when the detected workspace contains multiple nearby items.

Second, verification frequency determines when errors are detected. Although a full action checker increases execution time, it can correct mistakes by detecting failures early. In stacking, disabling verification leads to zero task success because early errors propagate and the final stack violates the target order. Using only a goal checker delays error detection until task completion: the system continues stacking despite previous mistakes, producing an invalid intermediate stack and requiring replanning from a more complex state, which results in unnecessary actions and increased execution time. Full action checker instead enables early detection and timely recovery.

Overall, these results point to a practical guideline: sparse verification improves evaluation efficiency, whereas dense verification is required when tasks involve strong dependencies or active disturbances.

\subsection{Ablation Experiment Protocol}

For the action-checker ablation, we evaluate performance using the task progress score in Appendix~\ref{app:task_setup_metrics} and measure efficiency by runtime under controlled human disturbances. We compare three action-checker configurations: no action checker, a goal checker only, and the full action checker. The goal-checker-only configuration performs a single verification at the end of execution, such as checking whether the final goal is satisfied after all planned actions complete. In contrast, the full action checker verifies outcomes step by step after each action.

We use the following disturbance protocols.

1. No disturbance, sorting. We execute the sorting task without human intervention, placing food items into the box and toys into the bowl.

2. One disturbance, sorting. During the first pick, we move the target object to induce a misgrasp and simulate grasp failure.

3. Two disturbances, sorting. In addition to the first-pick disturbance, we move the target placement location during the final place action to simulate misplacement under a shifted goal configuration.

4. One disturbance, stacking. To test a sequential task where later outcomes depend on earlier actions, we move the target block during the first grasp to induce a disturbance.

\subsection{VLA Fine-tuning Setup}
\label{app:vla_setup}

We fine-tune the pretrained $\pi_{0.5}$ VLA policy \cite{intelligence2025pi_} on two tabletop manipulation tasks: {sorting} and {stacking}. Training configurations are provided in Table~\ref{tab:vla_hyperparams}.

For sorting, we collect 40 human demonstrations, evenly split into two instruction variants: 
20 episodes of ``Grab the food to the bowl" and 20 episodes of ``Grab the toys to the bin". 
For stacking, we collected 30 demonstrations consisting of 10 episodes for each of three stacking orders: green-yellow-orange, yellow-green-orange, and orange-yellow-green.

We initialize the policy from the $\pi_{0.5}$-base checkpoint. 
To retain pretrained visual representations, we freeze the vision encoder and fine-tune only the reasoning and action components via Low-Rank Adaptation (LoRA) \cite{hu2022lora}. 
Specifically, we apply LoRA to the linear projection layers in the Attention and FFN modules for both the VLM backbone and the flow-matching action expert. For all methods, we use the same perception setup consisting of one global camera and one wrist camera to ensure a fair comparison.

\begin{table}[t]
\centering
\small
\setlength{\tabcolsep}{8pt}
\renewcommand{\arraystretch}{1.15}
\caption{VLA fine-tuning configuration. We apply LoRA to attention and FFN blocks for both the VLM backbone and the flow-matching action expert.}
\label{tab:vla_hyperparams}
\begin{tabularx}{\linewidth}{l X}
\toprule
\textbf{Component} & \textbf{Setting} \\
\midrule
Initialization & $\pi_{0.5}$-Base \\
\midrule
\textbf{VLM Backbone} & Gemma-2B \\
\hspace{2mm} Vision Encoder & Frozen \\
\hspace{2mm} LoRA Target & Attention \& FFN Linear Projections \\
\hspace{2mm} LoRA $(r,\alpha)$ & $(16, 16)$ \\
\midrule
\textbf{Action Expert} & Gemma-300M \\
\hspace{2mm} LoRA Target & Attention \& FFN Linear Projections \\
\hspace{2mm} LoRA $(r,\alpha)$ & $(32, 32)$ \\
\midrule
Training Steps & 10{,}000 \\
Batch Size & 64 \\
Hardware & 1$\times$ NVIDIA A6000 (48\,GB) \\
Training Duration & $\sim$38 Hours \\
\bottomrule
\end{tabularx}
\end{table}

\begin{figure}[H]
  \centering
  \includegraphics[width=\linewidth]{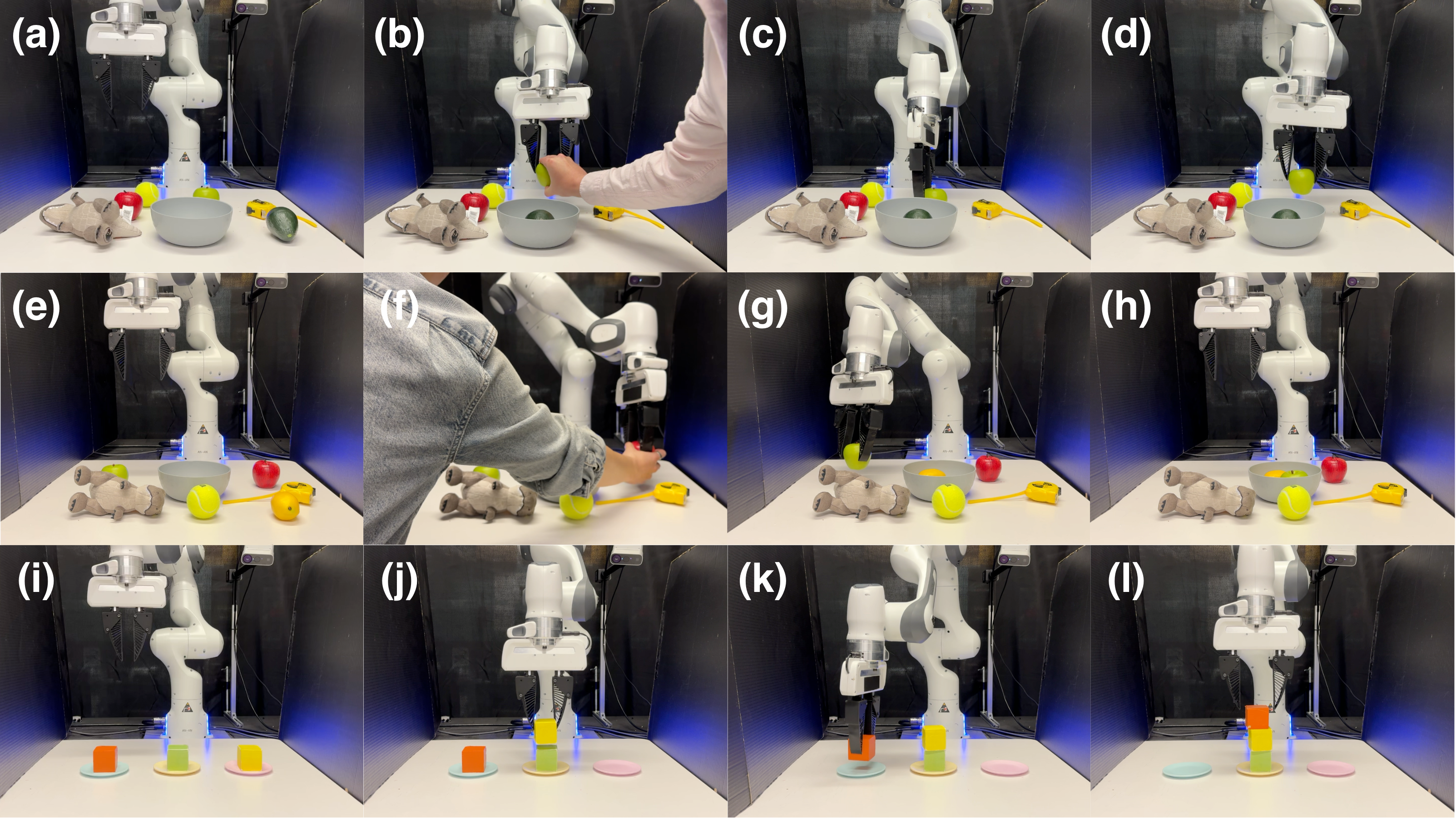}
  \caption{
  Comparison between PLanAR and TiPToP on the Franka Panda.
  (a--d) PLanAR on disturbed fruit sorting. After human disturbance during the second grasp, PLanAR detects the failed effect, retries the grasp, detects a second failed grasp caused by an inaccurate grasp pose, and replans again to complete the task.
  (e--h) TiPToP on the same disturbed fruit-sorting setup. Although TiPToP generates a sorting plan through cuTAMP, it does not detect the disturbed second grasp and terminates with only two fruits sorted.
  (i--l) PLanAR transfers to block stacking on the Franka Panda, while TiPToP does not generate an executable stacking plan in our setup.
  }
  \label{fig:tiptop}
\end{figure}

\subsection{Comparison of PLanAR and TiPToP}
\label{app:tiptop}

Fig.~\ref{fig:tiptop} provides a comparison between PLanAR and TiPToP on the Franka Panda. In the fruit-sorting task, PLanAR successfully transfers the same high-level reasoning, verification, and replanning logic to the Franka platform. As shown in Fig.~\ref{fig:tiptop}(a--d), we introduce a human disturbance during the second grasp by removing the grasped fruit and returning it to the table. PLanAR detects that the expected grasp effect is not satisfied and retries the grasp. Although the first retry also fails due to an inaccurate grasp pose, PLanAR detects this failure again, replans a second retry, and eventually completes the task by sorting all three fruits.

In contrast, TiPToP can generate a fruit-sorting plan with cuTAMP, as shown in Fig.~\ref{fig:tiptop}(e--h), but it does not detect the disturbed second grasp. As a result, it continues execution and terminates with only two fruits sorted. We further evaluate block stacking in Fig.~\ref{fig:tiptop}(i--l). PLanAR completes the stacking task on the Franka Panda using the same planning-language-grounded interface, whereas TiPToP does not generate an executable stacking plan with its motion planner in our setup. These examples illustrate that PLanAR's explicit verification and replanning are useful under disturbances and failed grasps, and that its primitive-based execution interface can provide a practical transfer path across robot embodiments.

\subsection{In-the-Wild Deployment on UR5e}
\label{app:wild_deployment}

\begin{figure}[H]
  \centering
  \includegraphics[width=\linewidth]{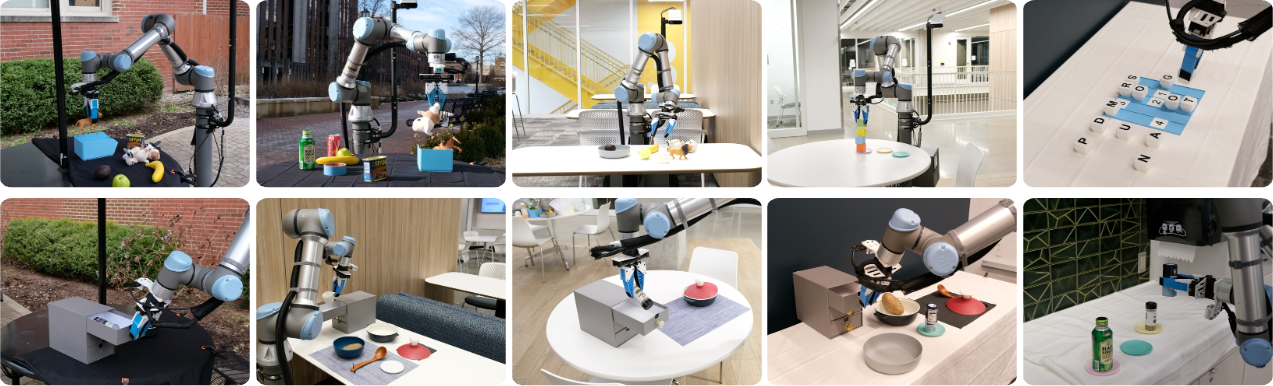}
  \caption{Qualitative in-the-wild deployment of PLanAR on the UR5e platform across diverse scenes.}
  \label{fig:wild}
\end{figure}

To evaluate PLanAR beyond controlled tabletop settings, we deploy the UR5e platform across diverse real-world scenes, including laboratory, indoor, and outdoor environments. As shown in Fig.~\ref{fig:wild}, these deployments qualitatively demonstrate that the platform can execute long-horizon manipulation tasks under varied backgrounds, lighting conditions, clutter, and object arrangements. They also provide real-world rollout data used for module-level evaluation, including visual grounding, action checking, grasp planning, and goal checking.

\end{document}